\documentclass[sigconf,natbib=false]{acmart}
\AtBeginDocument{%
  }

\copyrightyear{2025}
\acmYear{2025}
\setcctype{by}
\acmConference[KDD '25]{Proceedings of the 31st ACM SIGKDD Conference on Knowledge Discovery and Data Mining V.2}{August 3--7, 2025}{Toronto, ON, Canada}
\acmBooktitle{Proceedings of the 31st ACM SIGKDD Conference on Knowledge Discovery and Data Mining V.2 (KDD '25), August 3--7, 2025, Toronto, ON, Canada}
\acmDOI{10.1145/3711896.3737272}
\acmISBN{979-8-4007-1454-2/2025/08}

\RequirePackage[
  datamodel=acmdatamodel,
  style=acmnumeric,
  ]{biblatex}

\usepackage{subfigure}
\usepackage{graphicx}
\usepackage{colortbl}
\definecolor{inteins}{RGB}{128,179,255}
\definecolor{intzwei}{RGB}{42,127,255}
\definecolor{intdrei}{RGB}{0,85,212}
\definecolor{intvier}{RGB}{255,143,143}
\usepackage{algorithm}
\usepackage{algpseudocode}
\usepackage{amsmath}
\usepackage{color}
\usepackage{multirow}
\usepackage{algorithm}
\usepackage{algpseudocode}
\usepackage{subcaption}  
\usepackage{amsmath}
\usepackage{color}
\usepackage{multirow}
\begin{document}

\title{UoMo: A Universal Model of Mobile Traffic Forecasting for Wireless Network Optimization}

\author{Haoye Chai}
\affiliation{%
  \institution{Department of Electronic
Engineering, BNRist, \\ Tsinghua University}
  \country{Beijing, China}
  }
\email{haoyechai@mail.tsinghua.edu.cn}

\author{Shiyuan Zhang}
\affiliation{%
  \institution{Department of Electronic
Engineering, BNRist, \\ Tsinghua University}
  \country{Beijing, China}
  }
\email{zhangshi22@mails.tsinghua.edu.cn}

\author{Xiaoqian Qi}
\affiliation{%
  \institution{Department of Electronic
Engineering, BNRist, \\ Tsinghua University}
  \country{Beijing, China}
  }
\email{qixiaoqian24@mails.tsinghua.edu.cn}

\author{Baohua Qiu}
\affiliation{%
  \institution{China Mobile}
  \country{Beijing, China}
  }
\email{qiubaohua@chinamobile.com}

\author{Yong Li}
\affiliation{%
  \institution{Department of Electronic
Engineering, BNRist, \\ Tsinghua University}
  \country{Beijing, China}
  }
\email{liyong07@tsinghua.edu.cn}

\renewcommand{\shortauthors}{Haoye Chai et al.}

\begin{abstract}
Mobile traffic forecasting allows operators to anticipate network dynamics and performance in advance, offering substantial potential for enhancing service quality and improving user experience.
It involves multiple tasks, including long-term prediction, short-term prediction, and generation tasks that do not rely on historical data. By leveraging the different types of mobile network data generated from these tasks, operators can perform a variety of network optimizations and planning activities, such as base station (BS) deployment, resource allocation, energy optimization, \emph{etc.} However, existing models are often designed for specific tasks and trained with specialized data, 
and there is a lack of universal models for traffic forecasting across different urban environments.
In this paper, we propose a \underline{U}niversal m\underline{o}del for \underline{Mo}bile traffic forecasting (UoMo), aiming to handle diverse forecasting tasks of short/long-term predictions and distribution generation across multiple cities to support network planning and optimization. UoMo combines diffusion models and transformers, where various spatio-temporal masks are proposed to enable UoMo to learn intrinsic features of different tasks, and a contrastive learning strategy is developed to capture the correlations between mobile traffic and urban contexts, thereby improving its transfer learning capability.
Extensive evaluations on 9 real-world datasets demonstrate that UoMo outperforms current models in various forecasting tasks and zero/few-shot learning. It shows an average accuracy improvement of 27.85\%, 18.57\%, and 15.6\% in long-term prediction, short-term prediction, and generation tasks, respectively, showcasing its strong forecasting capability.
We deploy UoMo on China Mobile's JiuTian platform, leveraging the predicted mobile data to optimize live networks. This optimization includes BS deployment, resulting in a 25.3\% increase in served users, and BS sleep control, which reduces equipment depreciation by 40.7\%.
The source code is available online: \url{https://github.com/tsinghua-fib-lab/UoMo}.
\end{abstract}



\maketitle

\section{Introduction}

In recent years, foundation models~\cite{brown2020language, touvron2023llama, radford2021learning} have made substantial strides in natural language processing and computer vision. These models are reshaping the AI ecosystem by harnessing their powerful data processing, generalization, and zero/few-shot learning capabilities. 
An increasing number of specialized domains have developed foundational models tailored to their specific data and contextual demands, including healthcare, medicine, urban navigation, and beyond~\cite{10.1145/3636534.3649361, 10.1145/3583780.3615155, 10.1145/3630744.3658614, 10.1145/3641519.3657509}.
Mobile networks encompass massive amounts of mobile traffic, user, and geographical data, providing inherent data support for building universal models. However, such dedicated models for mobile network domains have yet to be established.
We hence aim to construct a universal model for mobile traffic forecasting,  which can handle rich features of large-scale mobile data while retaining the generalization needed across multiple network applications~\cite{10.1109/TNET.2021.3136707, 10.1145/3637528.3671544, 10.1145/3544216.3544251}.

Constructing such a universal model of mobile traffic forecasting is vital for mobile networks~\cite{10.1145/3637528.3671730, 10.1109/TNET.2021.3053771, 9714211}. On the one hand, mobile traffic forecasting offers great potential for network planning and optimization. It enables operators to anticipate traffic dynamics, facilitating proactive perceptions of resource utilization and service quality, and allowing for the preemptive development of optimization strategies.
On the other hand, mobile networks encompass a variety of optimization scenarios, including radio resource scheduling~\cite{6168145, 7314993}, Base Station (BS) deployment~\cite{9204664, 9456090}, and antenna configuration~\cite{10086045, 9664425}, \emph{etc}. These scenarios involve diverse objectives like throughput, coverage, and energy efficiency, which \emph{impose distinct tasks on traffic forecasting}. 
For example, radio resource scheduling requires performing short-term traffic forecasting tasks, prioritizing traffic dynamics to improve user experience~\cite{8896929, 10239288}, whereas BS deployment involves long-term traffic forecasting tasks, focusing on long-term traffic patterns within a region to align with network demands~\cite{9586045, 9140397}.
For the planning and deployment of live wireless networks, it is essential to leverage the powerful data mining and robust generalization capabilities of universal models to simultaneously address a variety of optimization tasks.


Although numerous notable works have emerged in the area of mobile traffic forecasting~\cite{10.1145/3589132.3625569, 10.1145/3643832.3661437, 10449458, 10528242, 10.1145/3485983.3494844, 10.1145/3580305.3599801, sheng2025unveilingpowernoisepriors},
current methods typically employ one-to-one approaches: designing customized models by leveraging task-specific data~\cite{8641378, 9919315, 9714211, 10.1145/3664655}.
The complex, customized models deployed in live networks often require manual orchestration and scheduling, which will lead to the waste of computational and storage resources, increasing the overhead of model deployment.
In addition, mobile traffic is inherently heterogeneous of various collection granularity and scope. For example, Measure Report (MR) data primarily collects millisecond-level user traffic, while Performance Management (PM) data gathers cell-level traffic statistics over 15-minute intervals~\cite{7299258, 8932276}, leading to the absence of a unified representation akin to that found in natural language. Consequently, it is challenging to directly apply pre-trained models from the natural language/visual domains to mobile traffic data. Although some efforts have been made to reprogram mobile traffic data into a natural language format~\cite{NEURIPS2023_3eb7ca52, jin2024timellmtimeseriesforecasting}, this approach heavily relies on the quality of manually crafted prompts, making it difficult to capture a universal representation of mobile traffic.
Specifically, current mobile traffic forecasting models face two key limitations:

i) \emph{Limited generalization}. Mobile traffic data is inherently shaped by the spatio-temporal dynamics of population distribution and communication demands. Due to variations in geographic environments, lifestyle habits, and urban layouts across different cities, mobile traffic can differ significantly~\cite{8984314, 10.1109/TNET.2016.2623950}. With relatively small parameters, current models struggle to capture the diverse spatio-temporal patterns inherent in large-scale data across multiple cities. Additionally, it is challenging to encapsulate the complex correlations between contextual factors and mobile traffic, resulting in poor transferability in multi-city scenarios.

ii) \emph{Constrained task adaptability}. Mobile traffic forecasting is extensively applied across varying optimization scenarios. However, current models are often designed with specialized modules tailored to specific tasks. For instance, in short-term forecasting, models usually focus on capturing traffic fluctuations that employ autoregressive or event-driven methods. In contrast, long-term predictions emphasize the regular patterns of traffic and typically utilize time series decomposition techniques. These dedicated models increase design complexity and raise deployment costs when applied to diverse scenarios.


To tackle the limitations, we propose a \underline{U}niversal m\underline{o}del for \underline{Mo}bile traffic forecasting (UoMo), which aims to learn universal features of mobile traffic data and to handle multiple tasks in mobile networks, thereby establishing a one-for-all forecasting model.
\textbf{First}, inspired by Sora~\cite{videoworldsimulators2024}, UoMo adopts the transformer-based diffusion model as the backbone instead of the U-Net structure, to help the model understand the diverse features of massive mobile data. 
We propose a contrastive diffusion algorithm and adjust the variational lower bound by analyzing the cross-entropy between mobile traffic and contextual features. This helps the model better integrate environmental information, improving generalization and addressing the first limitation.
\textbf{Second}, we adopt a task-oriented masking and self-supervised training paradigm, where we categorize traffic forecasting in mobile networks into three tasks: short-term prediction, long-term prediction, and generation. 
We design the corresponding masking strategies to enable the model to learn data features for various tasks and adapt to multiple tasks, thus addressing the second research challenge.

$\bullet$ To the best of our knowledge, it is the first universal model designed for mobile traffic forecasting.
The proposed model enables various forecasting tasks in mobile networks across different urban environments via a unified framework, assisting network operators in achieving highly efficient network planning and optimization.

$\bullet$ We develop our universal model using a masked diffusion approach with spatio-temporal masking strategies tailored for diverse forecasting tasks, including short/long-term predictions and distribution generation.
To strengthen the correlation between contextual features and mobile traffic, we further propose a context-aware contrastive learning fine-tuning strategy, which can enhance forecasting and transfer learning capabilities.

$\bullet$ We conduct extensive evaluations with 9 real-world mobile traffic datasets. The results validate UoMo's superior generalization, multi-task capabilities, and robust few/zero-shot performance in unseen scenarios.
We also deploy the UoMo model on China Mobile's Jiutian platform. Experiments on the live system data prove our model empowers multiple network optimization scenarios, including a 25.3\% increase in served users for BS deployment and a 40.7\% reduction in equipment depreciation for BS sleep control.

\section{Problem Formulation}
\label{sec:pre}

Mobile Traffic refers to the volume of data transmitted over wireless channels between mobile devices and BSs over a period of time.
We consider a discrete-time scenario $T=\{0,...T\}$ with equal time intervals. For a single BS, the traffic variation over time $T$ can be represented as $\{b_t\}_{t=0:T}$, where $b_t$ denotes the aggregated traffic within the coverage area of BS at time $t$.
To characterize the mobile traffic features across an urban region $\mathcal{G}$ with contextual features $C_\mathcal{G}$, we define the geographical length and width of that region as $H$ and $V$, respectively. The mobile traffic of region $\mathcal{G}$ can then be defined as the sum of the aggregate traffic of all BSs located at $\mathcal{G}$: $\mathcal{S}_{0:T} = \sum_b \{b_t\}_{t=0:T}$.
Regarding diverse tasks in communication networks, such as base station deployment, user access control, and wireless resource allocation, different traffic forecasting tasks are often required. We categorize forecasting into 3 typical tasks:

$\bullet$ \textbf{Short-term prediction task} uses long historical data $0 \sim t$ to predict mobile traffic dynamics over a short future period, $i.e.,$ $\{ \mathcal{S}_{0:t}, C_{\mathcal{G}} \} \rightarrow \mathcal{S}_{t:T}$, where $t \gg (T-t)$.
Based on the forecasts with fluctuations, operators can understand upcoming network demands and make optimizations such as resource allocation~\cite{9923420} and access control~\cite{10255286} to improve user experience in live networks.

$\bullet$ \textbf{Long-term prediction task} 
estimates future traffic patterns based on limited historical data, $i.e.,$ $\{ \mathcal{S}_{0:t}, C_{\mathcal{G}} \} \rightarrow \mathcal{S}_{t: T}$, where $t \ll T$. It explores inherent periodical patterns within the traffic. This forecasting enables operators to estimate and analyze network performance from a global perspective, thereby facilitating the formulation of network optimization planning strategies, such as cell dormancy~\cite{10390348} and network capacity expansion~\cite{10063388}.

$\bullet$ \textbf{Generation task} focuses on identifying underlying network demand within a specific area without referring to historical data, $i.e.,$ $\{ C_{\mathcal{G}} \} \rightarrow \mathcal{S}_{t=0:T}$.
It helps operators assess potential communication demands in new regions lacking historical data, allowing them to develop planning strategies such as BS deployment~\cite{10086045}, network segmentation~\cite{10634862}, and capacity planning~\cite{10164177}, \emph{etc}.

We aim to build a domain-universal model capable of achieving the above 3 forecast tasks. The problem can be defined as follows.

\emph{Problem Definition. Given an arbitrary urban region $\mathcal{G}$, the goal is to use a model $\mathcal{F}$ to forecast diverse mobile traffic sequence $\mathcal{S}$ with short/long/generation tasks, conditioning on urban contextual factors $C_{\mathcal{G}}$, i.e., $\mathcal{F}(\mathcal{S}_{t=T_0:T_1}/\mathcal{S}_{t=0:T_1}, C_{\mathcal{G}})$. }

However, building such a universal model is not straightforward. Specifically, two key challenges arise:
i). \textbf{What} strategies can be employed for the training process, ensuring that the model can handle the diverse forecasting tasks?
ii). \textbf{How} to effectively integrate user dynamics and contextual features with mobile traffic?

\section{Design of UoMo}
\label{sec:method}

\subsection{Framework overview}
\begin{figure}[ht]
  \centering
  \includegraphics[width=0.8\linewidth]{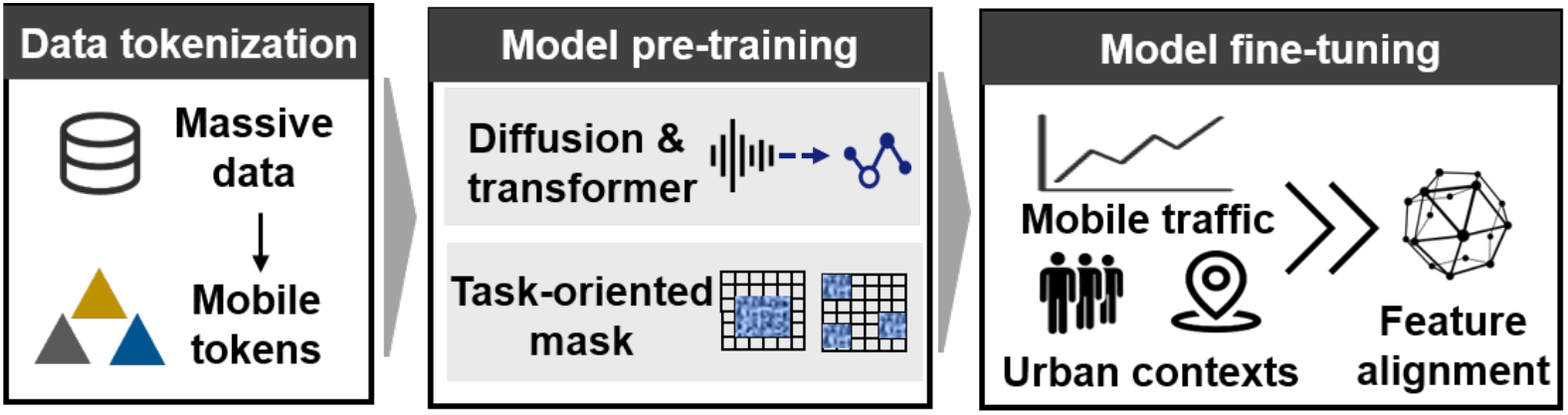}
  \caption{The flowchart of the UoMo framework}
  \label{frameowrk}
\end{figure}

To tackle the challenges, we propose the UoMo framework incorporating three stages as illustrated in Figure~\ref{frameowrk}.

i). \emph{Data tokenization} reshapes mobile traffic data from various spatial-temporal spans across multiple cities into a unified mobile token for model training and capturing their diverse features.

ii). \emph{Masked diffusion-based pre-training} tends to fully grasp the fundamental spatio-temporal features of mobile traffic across various forecasting tasks, where we design a diffusion-based backbone and task-oriented masks. 

iii). \emph{Urban context-aware fine-tuning} introduces a contrastive learning algorithm that integrates external factors closely associated with mobile traffic, including network user dynamics and urban POI distributions.

\subsection{Masked diffusion-based pre-training}

\begin{figure}[t]
  \centering
  \includegraphics[width=0.9\linewidth]{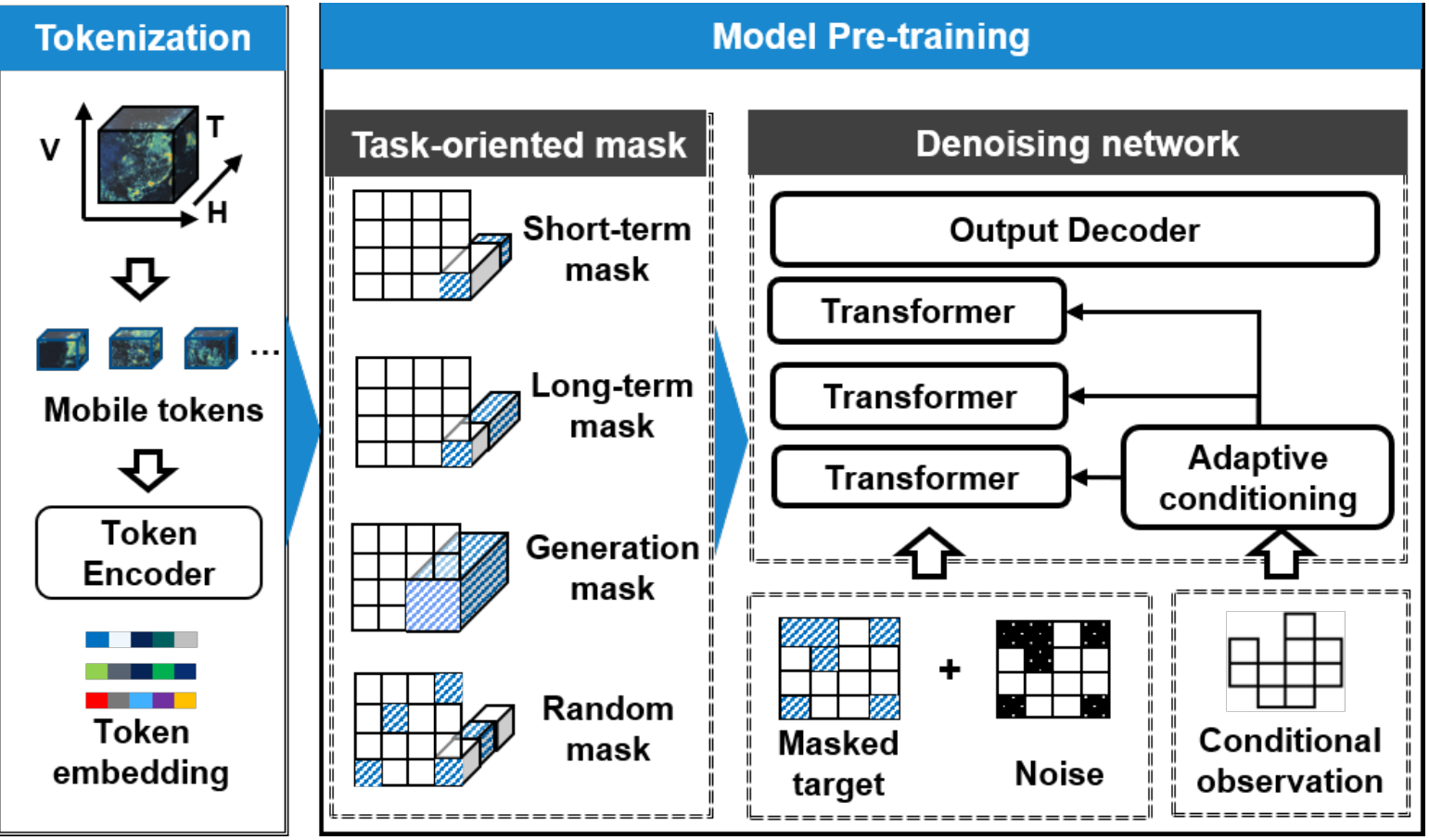}
  \caption{Masked diffusion-based pre-training network}
  \label{pretrain}
  \vspace{-0.5cm}
\end{figure}

We propose a masked diffusion model with self-supervised training, where specific masks are tailored for the three forecasting tasks to enhance the model's understanding of various forecasting tasks and to capture the spatio-temporal correlations inherent in massive mobile data, as shown in Figure~\ref{pretrain}.

\subsubsection{Mobile traffic data tokenization}

We draw inspiration from NLP tokenization, where we decompose traffic data with varying sampling intervals and diverse spatial ranges into basic unit $h_0 \times v_0 \times t_0$.
For traffic data $S$ of length $T$ within an urban region $H \times V$, the tokenization process breaks down $S$ into multiple small mobile tokens $X$, which can be expressed as
$S \in \mathbb{R}^{H\times V\times T} \rightarrow X \in \mathbb{R}^{(H'\times V'\times T') \times (h_0 \times v_0 \times t_0)},$
where $H' = H/h_0$, $V' = V/v_0$, and $T' = T/t_0$, the $(h_0, t_0, v_0)$ of X represents the mobile token. 
Subsequently, we use an embedding layer $E_x(X)$ ($e.g.,$ pooling layer, convolutional layer, or fully connected layer) to map the mobile token with hidden features $C$, $i.e.,$ $E_x(X) \in \mathbb{R}^{(H'\times V'\times T') \times C}$.

\subsubsection{Task-oriented mask}

After the mobile traffic data tokenization, the original $H \times V$ region is divided into multiple $h×v$ areas. The masking strategy performs self-supervised training by masking and reconstructing partial areas.
We define the masked parts as the target areas, and the unmasked parts around the target areas as the surrounding areas.
We develop 4 distinct masks: short-term, long-term, generation, and random masks, $m \in \mathbb{R}^{H'\times V'\times T'}$.
The first three focus on specific forecasting tasks, while random masking explores spatio-temporal correlations to enhance generalization.

$\bullet$ Short/Long-term masks. The schemes mask the time dimension $T'$ at a specific spatial location $(h,v)$ to reconstruct the mobile traffic within the period $T'-t_0$, where $t_0 \in \{0, T'\}$.
Depending on the ratio of $t_0$ to $T'$, the schemes correspond to short/long-term predictions:
\begin{equation}
    m_{h,v,t} = \{ 0, \ t_0 < t \leq T' \ | \ 1, \ 0 < t \leq t_0\}.
    \label{short_long_pre_mask}
\end{equation}

$\bullet$ Generation mask.
The generation mask completely obscures the temporal dimension at a specific spatial location $(h,v)$, enabling the model to generate mobile traffic sequence within the target area. Unlike prediction masks that rely on historical data, it captures spatio-temporal dependencies between the target area and its surrounding areas to generate underlying distributions:

\begin{equation}
    m_{h,v,t} = \{ 0, \ 0 \leq t \leq T' \}.
    \label{generation_mask}
\end{equation}

$\bullet$ Random mask.
It masks mobile traffic across both spatial and temporal dimensions, which aims to capture diverse correlations of mobile tokens, aiding the model in understanding the complex features of mobile data. Denote $\mathcal{R}(H', V', T')$  as randomly choosing items from $H'$, $V'$, and $T'$:
\begin{equation}
    m_{h,v,t} = \{0, \ \mathcal{R}(H', V', T') \ | \ 1, \ else  \}.
\end{equation}

\subsubsection{Self-supervised masked diffusion model}
After completing the multi-task masking process, the original mobile token $E_x(X)$ is divided into two parts: the masked portion $e$ requiring to be reconstructed, and the unmasked observation $o$:
\begin{equation}
  e = E_x(X) \odot m, \quad o = E_x(X) \odot (1-m),
\end{equation}
where $m$ corresponds to the four mask strategies, and $\odot$ represents element-wise products.
Subsequently, $o$ is input as conditions into the denoising network, while $e$ adds noise according to the forward process, which is given by
\begin{equation}
    e_k = \sqrt{\hat{\alpha}_k}e + (1-\hat{\alpha}_k) \epsilon, \ \ \epsilon \sim N(0, 1).
    \label{forward}
\end{equation}
Afterward, $e_k$ is fed into the transformer-based denoising network for further feature extraction.
To fully capture the dependencies between conditional observation and mobile traffic, we employ an adaptive conditioning method~\cite{10377858}. The method reshapes the scale and shift parameters of the layernorm of transformers by referring to the given conditions, which is proven to offer better effectiveness and computational efficiency~\cite{10.5555/3504035.3504518}. It can be formulated as follows:
\begin{equation}
\begin{aligned}
    \alpha, \beta, \gamma = \mathcal{F}_{\theta}(o), \quad e_k \leftarrow e_k +\alpha \mathcal{A}_{\theta}(\beta e_k +\gamma),
\end{aligned}
\end{equation}
where $\mathcal{F}_{\theta}$ and $\mathcal{A}_{\theta}$ denote linear layer and attention layer. $\alpha, \beta, \gamma$ are residual, scale, and shift parameters, respectively.
The denoising network aims to fit the posterior distribution of the diffusion process to predict the mean noise, ultimately reconstructing the final network traffic through an output decoder.
Our objective thus emphasizes the reconstruction accuracy of the masked portion, which can be given as indicated in (\ref{forward}):
\begin{equation}
    L_{\theta}= \underset{\theta }{min} \mathbb{E}_{e \sim q(e)}  \bigg \{ ||\epsilon - \epsilon_\theta(e_k,k|o)||^2 \odot m  \bigg \}.
    \label{lossfunction}
\end{equation}

\subsection{Urban context-aware fine-tuning}

\begin{figure}[t]
  \centering
  \includegraphics[width=\linewidth]{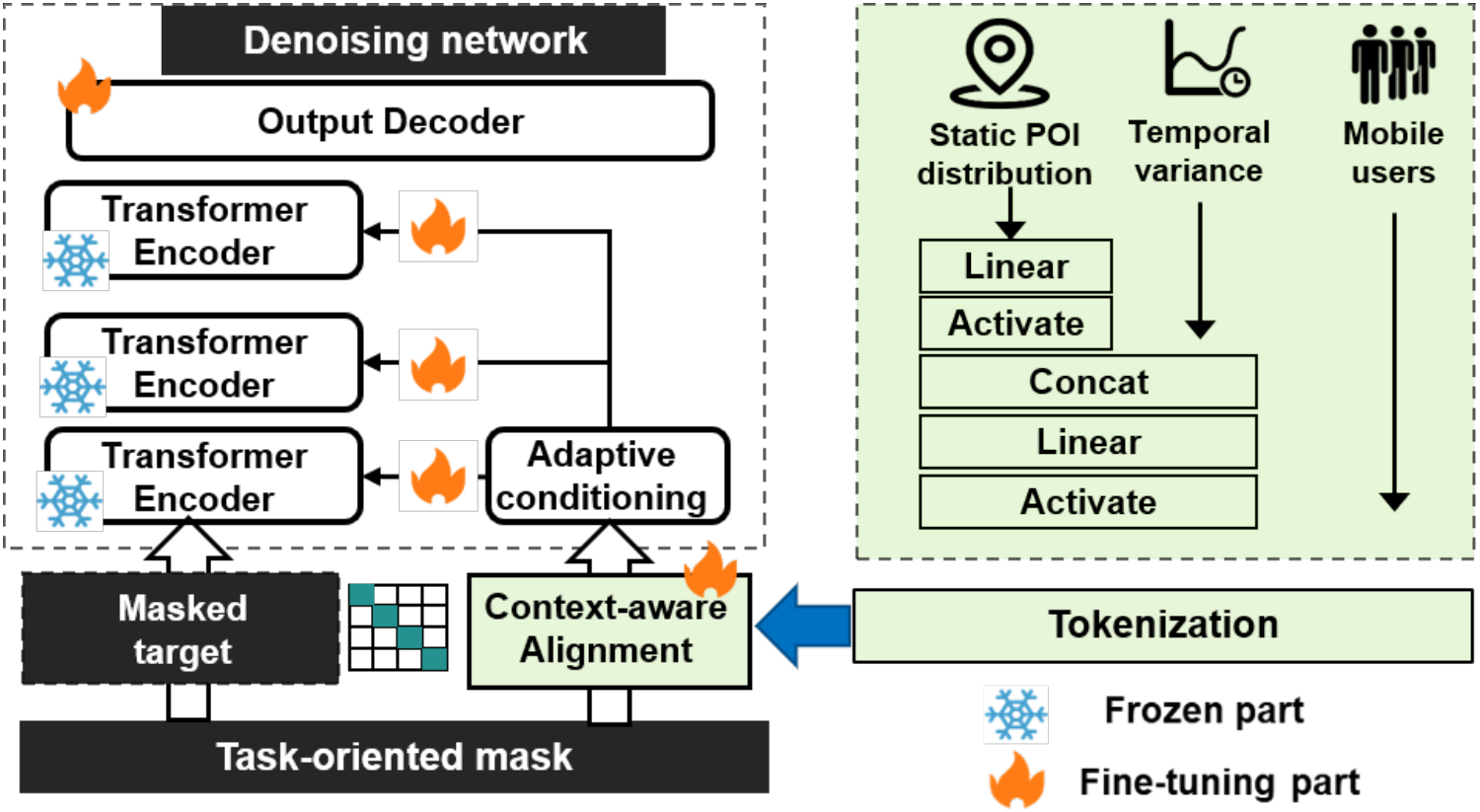}
  \caption{Context-aware fine-tuning process}
  \label{finetuning}
  \vspace{-0.5cm}
\end{figure}

Mobile traffic is not only a spatio-temporal sequence but also influenced by urban contexts. We thereby propose an urban context-aware fine-tuning scheme that integrates human dynamics and POIs into the UoMo, as shown in Fig~\ref{finetuning}.




\subsubsection{Contextual data transformation}

Mobile user refers to the number of users accessing the network, which can fully characterize the human dynamics in mobile networks.
Similar to mobile traffic, it is inherently a spatio-temporal sequence and denoted as $U \in \mathbb{R}^{H\times V\times T}$.  
We apply the same processing method as traffic tokens where we perform tokenization on mobile users as $c^u \in \mathbb{R}^{(H'\times V'\times T') \times (h_0 \times v_0 \times t_0)} $,
allowing this data to be directly input into the network for training.

POIs reflect the static distribution of urban layout.
For each base station's coverage area, we count the total number of each POI category within the coverage area so that we can obtain a POI vector $P \in \mathbb{R}^{H\times V} $.
Although the distribution of POIs is static, the impact of different categories of POIs on human behavior varies across different times, leading to corresponding variations in mobile traffic. For example, restaurant-type POIs typically show higher traffic during lunchtime and evening. In this regard, we design a dynamic POI transformation scheme. 
We first extract the intrinsic static features of POI distribution, which can be written as:
\begin{equation}
    h^s_p = \sigma (W^s\cdot P +B^s),
\end{equation}
where $\sigma$ is the Sigmoid activation function, $W^s$ and $B^s$ are the weight and bias parameters of MLP network. 
We further utilize an MLP network $\tau(t)$ to project timestamp as temporal embeddings,
where we use the nn.Embedding layer to encode the 2D-vector $t =[day, hour]$ and fuse the two embeddings using an MLP network, then we fuse the static POI feature with the temporal indicators:
\begin{equation}
    h^d_p = \sigma (W^l \cdot [h^s_p \oplus \tau(t)]  +B^l),
\end{equation}
where $\oplus$ denotes vector concatenation, $W^l$ and $B^l$ are the learnable parameters.
In this way, we can obtain spatio-temporal dynamic representations as $h_p^d \in \mathbb{R}^{H\times V\times T}$.
The final features of POI can be calculated via the same tokenization method as mobile traffic data: $c^p \in \mathbb{R}^{(H'\times V'\times T') \times (h_0 \times v_0 \times t_0)} $.
The ultimate contextual feature tokens can be denoted as $y=c^u+c^p$.

\subsubsection{Context-aware alignment}

To establish bridges between mobile traffic and contextual features,
we propose a contrastive learning algorithm. 
We define positive samples as mobile traffic tokens and contextual feature tokens within the same spatio-temporal block, denoted as $(e, y)$; while negative samples are defined as the two types of tokens from different spatio-temporal blocks, denoted
$(e', y)$.
Our goal is to maximize the mutual information between the traffic feature $e$ and contextual feature $y$. According to previous research~\cite{oord2019representationlearningcontrastivepredictive}, a density ratio can be utilized while preserving the mutual information as $I(e,y) \propto \frac{p(e|c)}{p(e)}$, and the maximization problem is equivalent to minimizing the InfoNCE loss that yields:
\begin{equation}
     min -\underset{e \in \mathbb{B}}{\mathbb{E}}log\frac{\frac{p(e|c)}{p(e)}}{\frac{p(e|c)}{p(e)}+\sum_{e'} \frac{p(e'|c)}{p(e')}} \geq log(N)-I(e,y),
\label{cl_loss2}
\end{equation}
where $\mathbb{B}$ denotes the entire batch of samples.

We claim in Lemma 1 that training the diffusion model with positive and negative samples is equivalent to minimizing the InfoNCE loss in contrastive learning of Eq~(\ref{cl_loss2}):

\emph{Lemma 1: By optimizing the Mean Squared Error (MSE) of positive and negative samples via Eq~(\ref{cl_loss_our}), we can achieve alignment between mobile traffic and contextual features.}
\begin{equation}
        L \approx \mathbb{E} \bigg \{ \bigg(\Vert \epsilon -\epsilon_{\theta}(e, k | y) \Vert^2 - \lambda \sum\nolimits_{e'} \Vert \epsilon -\epsilon_{\theta}(e',k | y) \Vert^2 \bigg) \odot m  \bigg \}.
\label{cl_loss_our}
\end{equation}

The proof of the lemma is provided in the appendix~\ref{proof_lemma}.
During the fine-tuning process, we partially froze the main parameters of the pre-trained model, including the attention layer, linear layer, and MLP network, to preserve the model's ability to learn general spatio-temporal features of mobile traffic. We primarily update the parameters in the adaptive conditioning and output decoder layers. By partially updating these components, the time and computational cost of the fine-tuning process can be reduced.

\section{Evaluation}
\label{sec:eval}

We perform evaluations on 9 real-world datasets to evaluate the UoMo with 13 baselines. The evaluations need to address the following 2 questions.

$\bullet$ \textbf{RQ1}: How does it perform in multi-task forecasting?

$\bullet$ \textbf{RQ2}: How does it perform in zero-shot and few-shot learning?

\subsection{Evaluation setting from live mobile system}

\subsubsection{Datasets}
\emph{Mobile traffic data.} We collect mobile traffic data of live networks from 7 cities of varying scales in China, encompassing downlink traffic including 4G and 5G data. The time granularity of the data ranges from 15 minutes to 1 hour.
Additionally, we utilize mobile traffic data from 2 other cities in China and Germany to validate UoMo's zero/few-shot capabilities.

\emph{Urban Contextual data.}
We collect mobile user data and mobile traffic data in each dataset.
We crawl POI data from each city through public map services, including 15 categories related to living, entertainment, and other aspects.

\begin{table*}[h]
\caption{Performance of short-term prediction task. Bold numbers denote the best results and $\underline{underline}$ numbers denote the second-best results. The proposed UoMo has the best prediction performance across 7 datasets.}
\label{shortterm}
{
\resizebox{0.88\textwidth}{!}{
\begin{tabular}{c|cc|cc|cc|cc|cc|cc|cc}
\hline
\textbf{Model}      & \multicolumn{2}{c|}{\textbf{Beijing}} & \multicolumn{2}{c|}{\textbf{Shanghai}} & \multicolumn{2}{c|}{\textbf{Nanjing}} & \multicolumn{2}{c|}{\textbf{Nanjing-4G}} & \multicolumn{2}{c|}{\textbf{Nanchang}} & \multicolumn{2}{c|}{\textbf{Nanchang-4G}} & \multicolumn{2}{c}{\textbf{Shandong}} \\ \hline
           & RMSE          & MAE          & RMSE          & MAE           & RMSE          & MAE          & RMSE           & MAE            & RMSE          & MAE           & RMSE            & MAE            & RMSE          & MAE          \\ \hline
\textbf{HA}         & 0.1199        & 0.0697       & 0.1151        & 0.0576        & 0.0788        & 0.0353       & 0.0830         & 0.0371         & 0.0589        & 0.0266        & 0.0702          & 0.0339         & 0.1739        & 0.0578       \\
\textbf{ARIMA}      & 0.2212        & 0.1333       & 0.1609        & 0.0819        & 0.1353        & 0.0622       & 0.1443         & 0.0668         & 0.1532        & 0.0666        & 0.1740          & 0.0789         & 0.1366        & 0.0531       \\ \hline
\textbf{SpectraGAN} & 0.2675        & 0.1228       & 0.2086        & 0.1226        & 0.2412        & 0.1186       & 0.2152         & 0.1151         & 0.2974        & 0.1467        & 0.1892          & 0.0935         & 0.2492        & 0.0814       \\
\textbf{keGAN}      & 0.3307         & 0.2994       & 0.3456        & 0.2174        & 0.3586        & 0.3318       & 0.3579         & 0.3297         & 0.3123        & 0.1913        & 0.2521          & 0.2206         & 0.2662        & 0.2616       \\
\textbf{Adaptive}   & 0.2779        & 0.2138       & 0.3007        & 0.2164        & 0.2606        & 0.1906       & 0.2219         & 0.1469         & 0.2305        & 0.1709        & 0.2572          & 0.1919         & 0.2688        & 0.1937       \\
\textbf{Open-Diff}    & 0.1104        & 0.0899       & 0.1326        & 0.0981        & 0.1087        & 0.0823       & 0.1196         & 0.1005         & 0.1204        & 0.0801        & 0.1377          & 0.0906         & 0.1166        & 0.0799       \\ \hline
\textbf{Time-LLM}   & 0.1511             & 0.1115            & 0.1388             & 0.0964             & 0.2351             & 0.1817            & 0.1754              & 0.1309              & 0.2039             & 0.1474             & 0.1770               & 0.1242              & 0.1571             & 0.0846            \\
\textbf{Tempo}      & 0.1206        & 0.0873       & 0.0747        & 0.0455        & 0.0805        & 0.0625       & 0.0652         & 0.0498         & 0.0830        & 0.0638        & 0.0749          & 0.0550         & 0.0969        & 0.0763       \\
\hline
\textbf{CSDI}       & 0.1752        & 0.1015       & 0.2060        & 0.1141        & 0.1722        & 0.0929       & 0.2299         & 0.1251         & 0.1797        & 0.0929        & 0.1587          & 0.0758         & 0.2131        & 0.0976       \\
\textbf{patchTST}   & 0.1107        & 0.0686       & 0.1288        & 0.0872        & 0.0935        & 0.0616       & 0.0960         & 0.0631         & 0.1182        & 0.0635        & 0.1162          & 0.0638         & 0.1089        & 0.0703      \\
\textbf{TimeGPT}    & 0.0598        & 0.0422       & 0.0866        & 0.0457        & 0.0646        & 0.0397       & 0.0657         & 0.0388         & 0.0502        & 0.0281        & $\underline{0.0576}$          & $\underline{0.0299}$         & 0.1219        & $\underline{0.0358}$       \\
\textbf{Lagllama}  & 0.0501        & 0.0349       & 0.0853        & $\underline{0.0441}$        & $\underline{0.0529}$        & $\underline{0.0302}$       & $\underline{0.0530}$         & $\underline{0.0286}$         & 0.0505        & 0.0271        & 0.0625          & 0.0322         & 0.1272        & 0.0371       \\
\textbf{UniST}      & $\underline{0.0332}$        & $\underline{0.0252}$       & $\underline{0.0658}$        & 0.0448        & 0.0623        & 0.0442       & 0.0608         & 0.0409         & $\underline{0.0433}$        & $\underline{0.0246}$        & 0.0852          & 0.0525         & $\underline{0.0766}$        & 0.0489       \\
\textbf{UoMo (our)}   & \textbf{0.0284} & \textbf{0.0135} & \textbf{0.0588} & \textbf{0.0349} & \textbf{0.0442} & \textbf{0.0247} & \textbf{0.0439} & \textbf{0.0143} & \textbf{0.0360} & \textbf{0.0178} & \textbf{0.0408} & \textbf{0.0221} & \textbf{0.0609} & \textbf{0.0343}      \\ \hline
\textbf{Improvement}   & \textbf{14.45\%} & \textbf{46.42\%} & \textbf{10.63\%} & \textbf{22.44\%} & \textbf{16.44\%} & \textbf{18.21\%} & \textbf{16.60\%} & \textbf{50.00\%} & \textbf{16.85\%} & \textbf{27.64\%} & \textbf{29.16\%} & \textbf{26.08\%} & \textbf{20.49\%} & \textbf{4.19\%}          \\ \hline
\end{tabular}
}
}
\end{table*}

\subsubsection{Baselines}
We select 13 baselines with 4 types.
i). Statistical models. Historical moving average (\textbf{HA}) and \textbf{ARIMA}~\cite{10.1145/3573900.3591123}.
ii) NLP-based model.
\textbf{Time-LLM}~\cite{jin2024timellmtimeseriesforecasting} and \textbf{Tempo}~\cite{cao2024tempopromtrained} describes time series features using natural language and uses these descriptions for forecasting.
iii). Spatio-temporal based models.
\textbf{CSDI}~\cite{NEURIPS2021_cfe8504b}, 
\textbf{TimeGPT}~\cite{garza2023timegpt1}, \textbf{Lagllama}~\cite{rasul2024lagllamafoundationmodelsprobabilistic}, \textbf{PatchTST}~\cite{nie2023timeseriesworth64}, and \textbf{UniST}~\cite{10.1145/3637528.3671662} forecast mobile traffic as spatio-temporal series via autoregression, decomposition, and spatial convolution. 
iv). Dedicated models for mobile networks.
\textbf{SpectraGAN}~\cite{10.1145/3485983.3494844}, \textbf{KEGAN}~\cite{10.1145/3580305.3599853}, 
\textbf{ADAPTIVE}~\cite{10.1145/3580305.3599801},  and \textbf{Open-Diff}~\cite{10.1145/3637528.3671544} utilizes dedicated contextual data to generate mobile traffic.
We describe the collected data and baselines in the appendix~\ref{datacollect_des} and \ref{bsl_des}.

\subsection{Multitask forecasting (RQ1)}

In our experiments, the temporal length is 64. For short-term prediction, the model forecasts 16 future points using the previous 48. For long-term prediction, the model forecasts 48 future points using the previous 16. For data generation, the model predicts all 64 points based on the current timestamp.
We primarily use the 35M UoMo model (with 16 transformer layers and a hidden feature size of 256), and the other scaling evaluations are provided in Table~\ref{running}.

\begin{figure}[t]
    \centering
    \subfigure[Forecasting performance of Beijing dataset.]{\includegraphics[width=\linewidth]{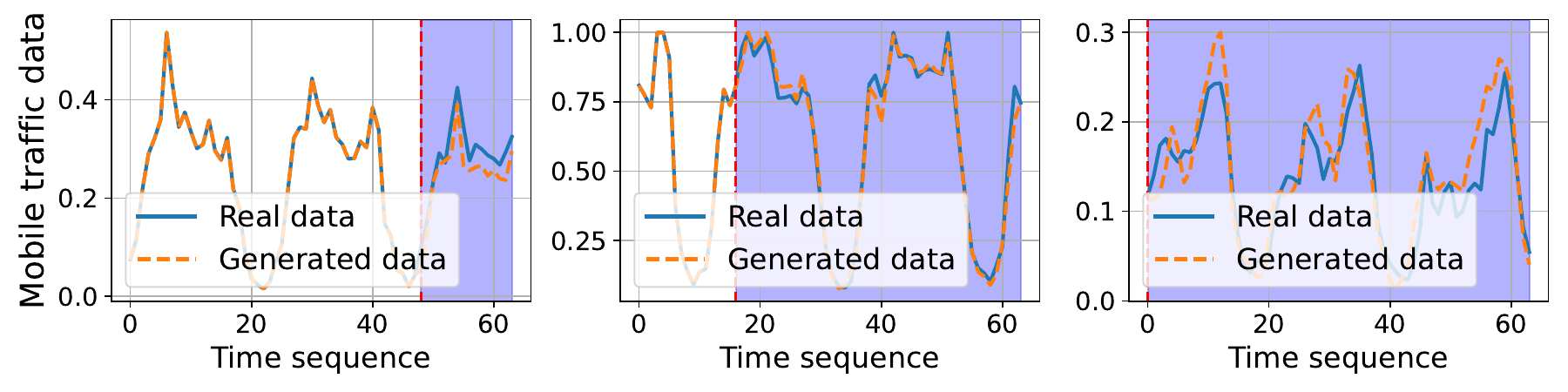}}
    \subfigure[Forecasting performance of Nanchang dataset.]{\includegraphics[width=\linewidth]{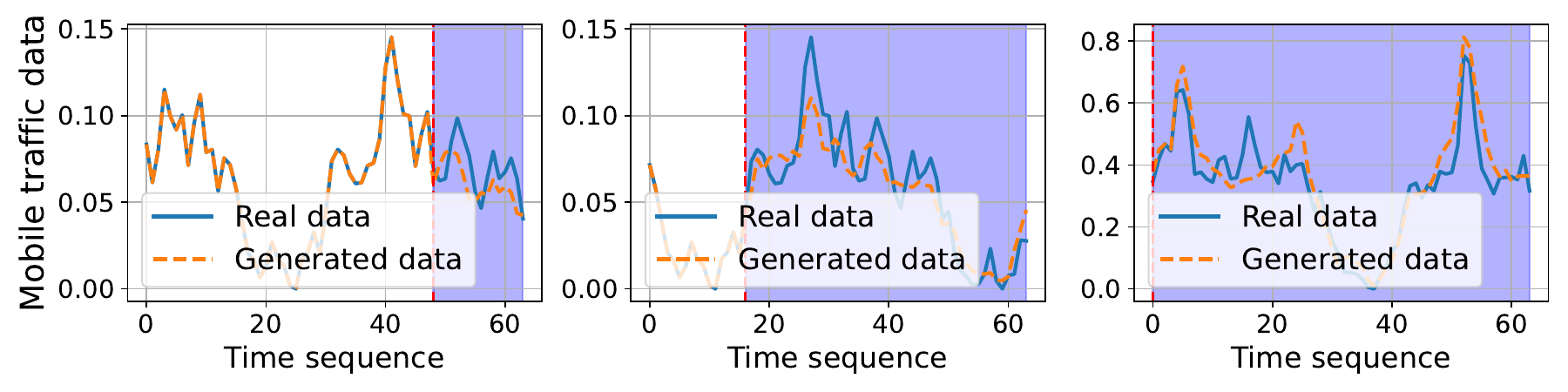}}
   
    \caption{Visualization results. From left to right, it represents short-term/long-term/generation tasks.}
    \label{visual_res}
    \vspace{-0.5cm}
\end{figure}

To intuitively demonstrate our model's universality for different tasks, we select two datasets as examples (Beijing and Nanchang) and plot the forecasting results in Figure~\ref{visual_res}. From left to right, it represents the tasks of short-term prediction $\rightarrow$ long-term prediction $\rightarrow$ traffic generation. The blue-shaded area indicates the model's predicted results, while the unshaded area represents historical observations. 
UoMo generates mobile traffic closely aligned with real values across all tasks, accurately predicting periodic trends and capturing fast dynamics, which shows that our UoMo model achieves forecasting across multiple cities and tasks, highlighting its generalization capability.

\begin{table*}[ht]
\caption{Performance of Long-term prediction task. }
\label{longterm}
{
\resizebox{0.88\textwidth}{!}{
\begin{tabular}{c|cc|cc|cc|cc|cc|cc|cc}
\hline
\textbf{Model}      & \multicolumn{2}{c|}{\textbf{Beijing}} & \multicolumn{2}{c|}{\textbf{Shanghai}} & \multicolumn{2}{c|}{\textbf{Nanjing}} & \multicolumn{2}{c|}{\textbf{Nanjing-4G}} & \multicolumn{2}{c|}{\textbf{Nanchang}} & \multicolumn{2}{c|}{\textbf{Nanchang-4G}} & \multicolumn{2}{c}{\textbf{Shandong}} \\ \hline
           & RMSE          & MAE          & RMSE          & MAE           & RMSE          & MAE          & RMSE           & MAE            & RMSE          & MAE           & RMSE            & MAE            & RMSE          & MAE          \\ \hline
\textbf{HA}         & 0.2945       & 0.1887       & 0.2214       & 0.1180        & 0.1808       & 0.0877       & 0.1914         & 0.0941        & 0.2011        & 0.0948       & 0.2285         & 0.1116         & 0.1331        & 0.0409       \\
\textbf{ARIMA}      & 0.2023       & 0.1237       & 0.1560       & 0.0811        & 0.1269       & $\underline{0.0592}$       & 0.1340         & $\underline{0.0634}$        & 0.1533        & $\underline{0.0709}$       & 0.1751         & 0.0848         & 0.1224        & 0.0380       \\ \hline
\textbf{SpectraGAN} & 0.3880       & 0.3005       & 0.1962       & 0.1234        & 0.3621       & 0.2717       & 0.3212         & 0.2160        & 0.2432        & 0.1787       & 0.2352         & 0.1260         & 0.2438        & 0.0809       \\
\textbf{keGAN}      & 0.3041       & 0.3716       & 0.2695       & 0.1837        & 0.2525       & 0.1809       & 0.2623         & 0.1917        & 0.2241        & 0.1770       & 0.2132         & 0.1837         & 0.1742        & 0.1315       \\
\textbf{Adaptive}   & 0.2885       & 0.2234       & 0.3019       & 0.2197        & 0.2631       & 0.1876       & 0.2619         & 0.1907        & 0.1959        & 0.1419       & 0.2436         & 0.1752         & 0.1605        & 0.1144       \\
\textbf{Open-Diff}    & 0.2801       & 0.1993       & 0.1562       & 0.1102        & 0.2042       & 0.1313       & 0.1809         & 0.1623        & 0.1822        & 0.1511       & 0.1907         & 0.1749         & 0.1496        & 0.1097       \\ \hline
\textbf{Time-LLM}   & 0.1472            & 0.1099            & 0.1765            & 0.1124             & 0.2463            & 0.1843            & 0.2239              & 0.1602             & 0.2261             & 0.1751           & 0.2199              & 0.1621              & 0.1789             & 0.0868            \\ 
\textbf{Tempo}  & 0.3514       & 0.2559       & 0.1518       & 0.0787        & 0.2896       & 0.1892       & 0.2780         & 0.1793        & 0.2380        & 0.1347       & 0.2365         & 0.1306         & 0.1020        & 0.0275       \\
\hline
\textbf{CSDI}       & 0.3822       & 0.2836       & 0.2880       & 0.1856        & 0.4164       & 0.3034       & 0.3492         & 0.2520        & 0.3879        & 0.2913       & 0.3452         & 0.2347         & 0.2973        & 0.1705       \\
\textbf{patchTST}   & 0.1512       & 0.1331       & 0.1627       & 0.0817        & 0.1521       & 0.1236       & 0.1644         & 0.0999        & 0.1430        & 0.0905       & 0.1789         & 0.1060         & 0.0985        & 0.0676       \\
\textbf{TimeGPT}    & 0.3422       & 0.2433       & $\underline{0.1110}$       & $\underline{0.0766}$        & 0.2272       & 0.1391       & 0.2116         & 0.1345        & 0.1994        & 0.1001       & 0.1953         & 0.0919         & 0.0887        & $\underline{0.0253}$       \\
\textbf{Lagllama}      & 0.2318       & 0.1879       & 0.1453       & 0.0874       & $\underline{0.0960}$      & 0.0683      & $\underline{0.1115}$         & 0.0959        & $\underline{0.1091}$        & 0.0788       & 0.1684         & 0.0889         & 0.1076        & 0.0439       \\
\textbf{UniST}      & $\underline{0.1426}$       & $\underline{0.1014}$       & 0.1264       & 0.0803        & 0.1831       & 0.0845       & 0.1268         & 0.0869        & 0.1387        & 0.0835       & $\underline{0.1445}$         & $\underline{0.0763}$         & $\underline{0.0622}$        & 0.0337       \\
\textbf{UoMo (our)}   & \textbf{0.1035} & \textbf{0.0696} & \textbf{0.0983} & \textbf{0.0679} & \textbf{0.0818} & \textbf{0.0532} & \textbf{0.0849} & \textbf{0.0570} & \textbf{0.0853} & \textbf{0.0576} & \textbf{0.1206} & \textbf{0.0563} & \textbf{0.0518} & \textbf{0.0197}      \\ \hline
\textbf{Improvement}   & \textbf{27.41\%} & \textbf{31.36\%} & \textbf{11.44\%} & \textbf{11.35\%} & \textbf{14.79\%} & \textbf{10.14\%} & \textbf{23.85\%} & \textbf{10.09\%} & \textbf{21.81\%} & \textbf{18.76\%} & \textbf{16.54\%} & \textbf{26.21\%} & \textbf{16.72\%} & \textbf{22.13\%}           \\ \hline
\end{tabular}
}
}
\end{table*}

\begin{table*}[ht]
\caption{Performance of generation task.}
\label{generation_res}
{
\resizebox{0.9\textwidth}{!}{
\begin{tabular}{c|cc|cc|cc|cc|cc|cc|cc}
\hline
\textbf{Model}      & \multicolumn{2}{c|}{\textbf{Beijing}} & \multicolumn{2}{c|}{\textbf{Shanghai}} & \multicolumn{2}{c|}{\textbf{Nanjing}} & \multicolumn{2}{c|}{\textbf{Nanjing-4G}} & \multicolumn{2}{c|}{\textbf{Nanchang}} & \multicolumn{2}{c|}{\textbf{Nanchang-4G}} & \multicolumn{2}{c}{\textbf{Shandong}} \\ \hline
                    & JSD               & MAE               & JSD                & MAE               & JSD               & MAE               & JSD                 & MAE                & JSD                & MAE               & JSD                 & MAE                 & JSD               & MAE               \\ \hline
\textbf{SpectraGAN} & 0.3621            & 0.1584            & 0.3788             & 0.1284            & 0.3477            & 0.2888            & 0.3352              & 0.2494             & $\underline{0.2285}$            & $\underline{0.1288}$            & 0.3482              & 0.1762              & 0.3364            & 0.1794            \\
\textbf{keGAN}      & 0.3435            & 0.2297            & 0.4909             & 0.2183            & 0.3071            & 0.1801            & 0.4862              & 0.1865             & 0.4032             & 0.1988            & 0.4792              & 0.2493              & 0.4007            & 0.2387            \\
\textbf{Adaptive}   & 0.3044            & 0.2143            & 0.2751             & 0.1848            & 0.3040            & 0.1536            & 0.2587              & 0.2304             & 0.2730             & 0.2251            & 0.3201              & 0.1681              & 0.2806            & 0.1889            \\
\textbf{CSDI}       & 0.3385            & 0.1431            &   0.2331          & 0.1025           & 0.4044            & 0.1844            & 0.3875              & 0.2516             & 0.3416             & 0.2163            & 0.2895              & 0.1734              & 0.2666            & 0.2170            \\
\textbf{Open-Diff}    & $\underline{0.2155}$            & $\underline{0.1112}$            & $\underline{0.2299}$             & $\underline{0.1020}$            & $\underline{0.2114}$            & $\underline{0.1149}$            & $\underline{0.2275}$              & $\underline{0.1322}$             & 0.2296            & 0.1326            & $\underline{0.2624}$              & $\underline{0.1222}$              & $\underline{0.1977}$            & $\underline{0.1203}$            \\
\textbf{UoMo (our)}   & \textbf{0.2013} & \textbf{0.0894} & \textbf{0.2259} & \textbf{0.1002} & \textbf{0.1971} & \textbf{0.0948} & \textbf{0.2164} & \textbf{0.0938} & \textbf{0.2226} & \textbf{0.1043} & \textbf{0.2494} & \textbf{0.1159} & \textbf{0.1558} & \textbf{0.0993}
      \\ \hline
\textbf{Improvement}   & \textbf{6.58\%} & \textbf{19.60\%} & \textbf{1.74\%} & \textbf{1.76\%} & \textbf{6.76\%} & \textbf{17.49\%} & \textbf{4.87\%} & \textbf{29.04\%} & \textbf{2.58\%} & \textbf{19.02\%} & \textbf{4.95\%} & \textbf{5.15\%} & \textbf{21.19\%} & \textbf{17.45\%}       \\ \hline
\end{tabular}
}
}
\end{table*}

$\bullet$ Short-term prediction.
The results are presented in Table~\ref{shortterm}. Since sufficient historical data is available for reference, most baselines, leveraging their temporal feature extraction modules, effectively predict short-term changes. 
Overall, UoMo improved the average RMSE performance by 17.80\% and the average MAE by 27.85\% across 7 datasets, where UoMo can improve the RMSE metric by up to 29.1\% (Nanchang-4G dataset) and the MAE metric by up to 50\% (Nanjing-4G dataset),
which exhibits stronger generalization capabilities compared to other models.
Through the adaptive layernorm module, the diffusion model iteratively integrates contextual features and leverages the transformer to capture long-term dependencies between mobile traffic and the environment. We believe this correlation can transfer across different cities, improving the model's generalization capability.

$\bullet$ Long-term prediction.
The results are also shown in Table~\ref{longterm}. For this task, the lack of sufficient historical observations often leads to performance degradation in some baselines. However, UoMo consistently achieves the best performance, 
It improves the average RMSE performance by 18.93\% and the average MAE by 18.57\% across 7 real-world datasets, with a maximum improvement of 31.36\% in MAE and a maximum enhancement of 27.41\% in RMSE (Beijing dataset), which showcases its adaptability to various tasks. 

$\bullet$ Mobile traffic generation.
As shown in Table~\ref{generation_res}, the absence of historical observation of the generation task prevents some existing baselines from completing the task. Nevertheless, the UoMo still achieved strong generative results. 
It achieves a 6.95\% improvement in the average JSD performance and a 15.6\% improvement in the average MAE across all the datasets, with up to 21.19\% improvement in the JSD metric (Shandong dataset), and a maximum enhancement of 29.04\% in MAE (Nanjing-4G dataset).
This is due to the contextual feature fusion module used during fine-tuning, which captures the correlation between contextual and mobile traffic features through contrastive learning. This allows the model to infer potential traffic distribution based on environmental changes, even without historical data.

\begin{figure}[t]
    \centering
    \subfigure[Results on Munich dataset.]{\includegraphics[width=0.45\linewidth]{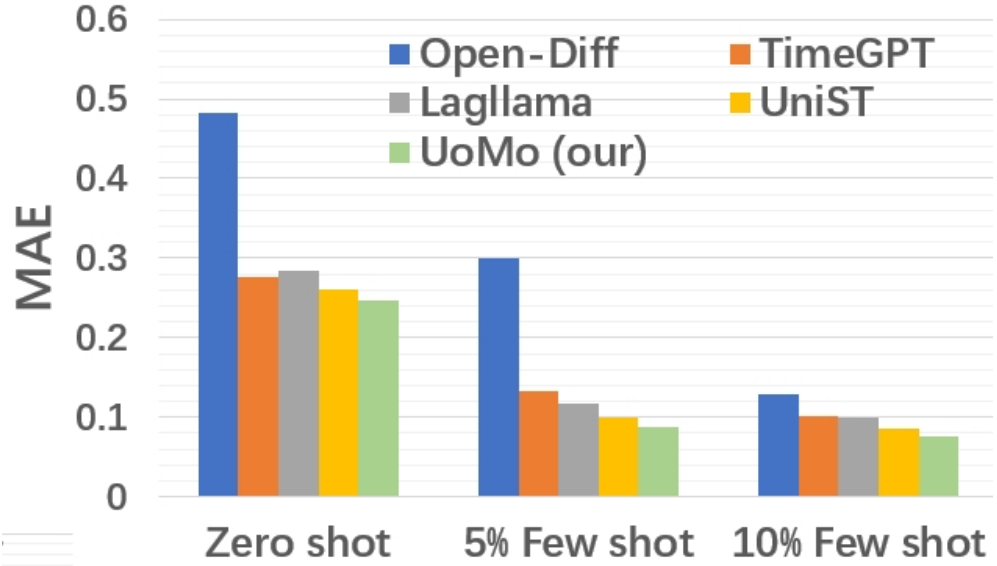}}
    \quad
    \subfigure[Results on Hangzhou dataset.]{\includegraphics[width=0.45\linewidth]{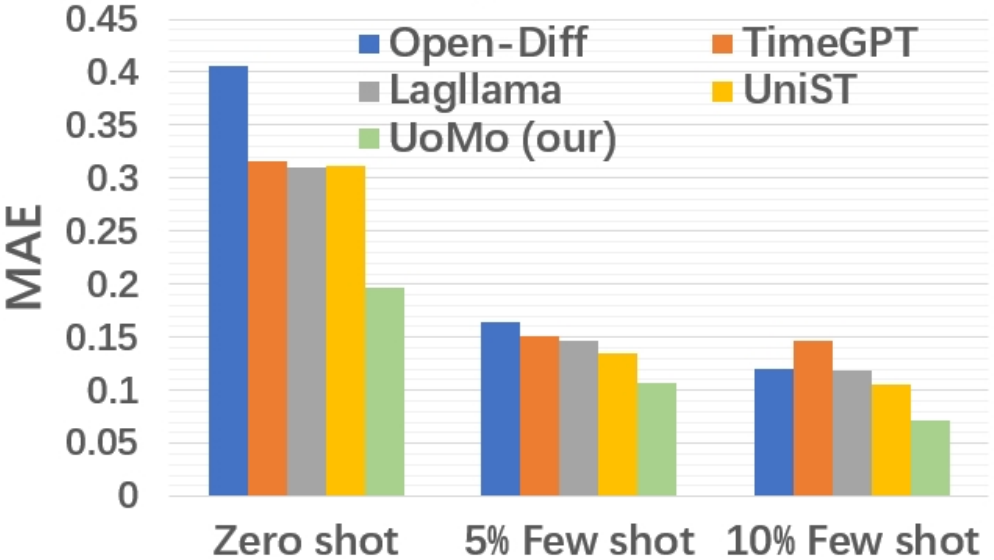}}
    \caption{Zero/Few-shot across two cities (Munich: long-term prediction task, Hangzhou: short-term prediction task).}
    \vspace{-0.5cm}
    \label{zero_few1}
\end{figure}

\begin{figure}[t]
    \centering
    \subfigure[Long-term prediction results of zero/few-shots on Munich dataset.]{\includegraphics[width=\linewidth]{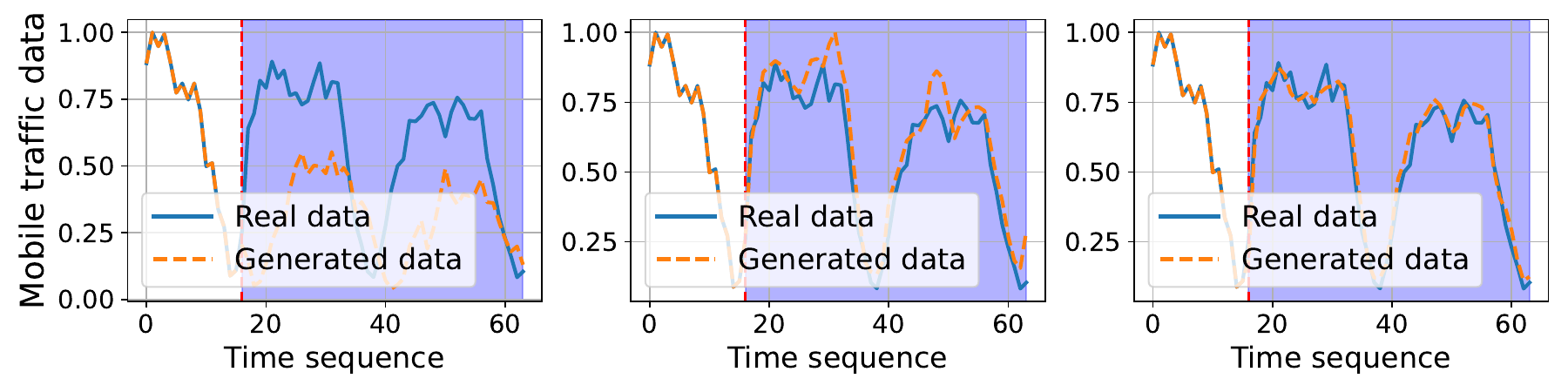}}
    \subfigure[Long-term prediction results of zero/few-shots on Hangzhou dataset.]{\includegraphics[width=\linewidth]{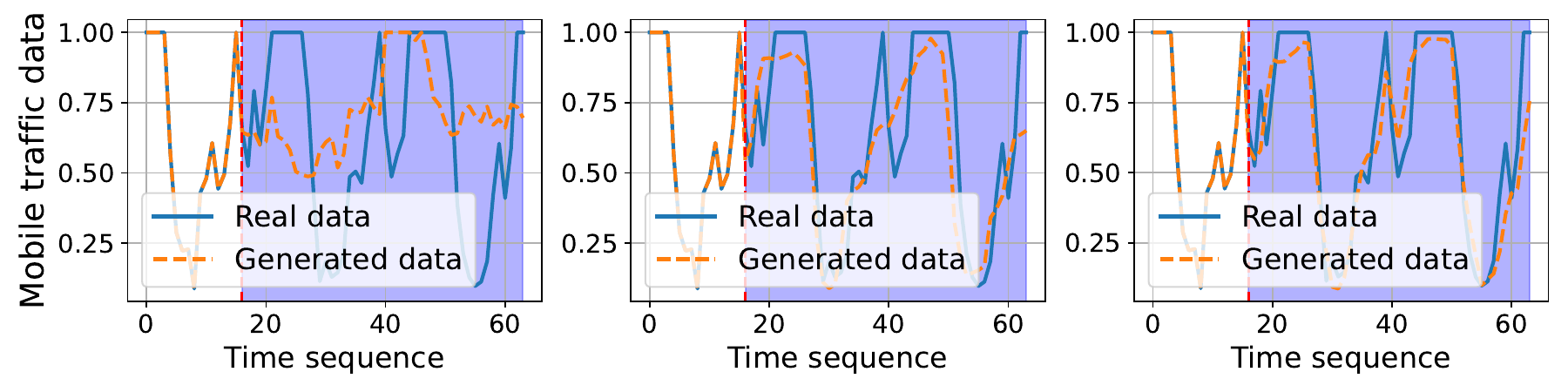}}
    \caption{Visualization of zero/few-shot learning. From left to right: zero-shot$\rightarrow$ 5\% few-shot$\rightarrow$ 10\% few-shot.}
    \label{zero_few2}
\end{figure}

\begin{table}[t]
\caption{Zero/Few-shot evaluation on FY and NN datasets.}
\label{zero_add}
{
\resizebox{0.4\textwidth}{!}{
\begin{tabular}{c|c|c|c}
\hline
\textbf{Model}              & \textbf{\begin{tabular}[c]{@{}c@{}}Zero-shot\\ (MAE)\end{tabular}} & \textbf{\begin{tabular}[c]{@{}c@{}}5\% Few-shot\\ (MAE)\end{tabular}} & \textbf{\begin{tabular}[c]{@{}c@{}}10\% Few-shot\\ (MAE)\end{tabular}} \\ \hline
\textbf{UoMo (FY)}      & 0.156                                                              & 0.034                                                                 & 0.023                                                                  \\ \hline
\textbf{LagLlama (FY)}  & 0.162                                                              & 0.094                                                                 & 0.053                                                                  \\ \hline
\textbf{UoMo (NN)}     & 0.134                                                              & 0.051                                                                 & 0.021                                                                  \\ \hline
\textbf{LagLlama (NN)} & 0.152                                                              & 0.088                                                                 & 0.026                                                                  \\ \hline
\end{tabular}
}
}
\end{table}

\subsection{Zero/Few shot learning (RQ2)}

To evaluate UoMo's zero/ few-shot learning capabilities, we select two datasets that UoMo has not encountered during training: Hangzhou (China) and Munich (Germany).
We choose 4 baselines that perform well in previous multitask forecasting: Open-Diff, TimeGPT, Lagllama, and UniST. The results are shown in Figure~\ref{zero_few1}, where 5\% few-shot and 10\% few-shot represent the model training with a small amount of data (5\% and 10\%, respectively). It shows that UoMo exhibits good zero-shot performance, especially in the Munich dataset, where UoMo’s zero-shot performance even surpasses that of Open-Diff after small-scale training.
After training with a small amount of data, all models show varying degrees of improvement. UoMo still demonstrates the best performance, indicating that UoMo can utilize the pre-trained model to quickly capture general features within unseen mobile data.
We visualize the performance in zero/few-shot scenarios, as shown in Figure~\ref{zero_few2}. We select a long-term forecasting task, with the results for zero-shot$\rightarrow$ 5\% few-shot$\rightarrow$ 10\% few-shot from left to right in the figure. It can be observed that UoMo can learn the general distribution characteristics of mobile traffic in the zero-shot phase, and after training with a small sample, the model realizes accurate traffic forecasting.
We additionally select two datasets, Fuyang (FY) and Nanning (NN), to evaluate the model. These datasets represent two Chinese cities of different scales, for which we collect mobile traffic data at an hourly granularity. As shown in Table~\ref{zero_add}, compared to the baseline method, UoMo is also capable of quickly learning the traffic patterns of new cities with limited training data, demonstrating its transferability.

\subsection{Ablation study}
\label{aba_sec}

To test the effectiveness of our proposed fine-tuning module, we conduct ablation experiments on UoMo, as shown in Table~\ref{ablation}, with UoMo-user and UoMo-POI representing the incorporation of mobile users and POI distributions, respectively, during the fine-tuning process.
It can be observed that adding these two contextual features during fine-tuning enhances model performance to varying degrees. Moreover, the performance degradation of UoMo-POI is more significant, indicating that mobile users better reflect the dynamic characteristics of mobile traffic and are more critical for mobile traffic forecasting compared to POI distribution.

\begin{table}[ht]
\caption{Performance of ablation study. $\Delta$ 
 represents the degradation after removing certain modules.}
\label{ablation}
{
\resizebox{0.5\textwidth}{!}{
\begin{tabular}{c|cc|cc|cc}
\hline
\textbf{Model}        & \multicolumn{2}{c|}{\textbf{Beijing}} & \multicolumn{2}{c|}{\textbf{Shanghai}} & \multicolumn{2}{c}{\textbf{Nanchang}} \\ \hline
\textbf{}             & Prediction        & Generation        & Prediction         & Generation        & Prediction        & Generation        \\
\multicolumn{1}{l|}{} & (RMSE)            & (JSD)             & (RMSE)             & (JSD)             & (RMSE)            & (JSD)             \\ \hline
\textbf{UoMo (our)}   & 0.1035            & 0.2213            & 0.0983             & 0.2202            & 0.0360            & 0.2226            \\ \hline
\textbf{UoMo w/o POI}    & 0.1230            & 0.2294            & 0.1295             & 0.2264            & 0.0421            & 0.2260            \\ \hline
$\Delta$        & -23.82\%          & -21.77\%          & -47.75\%           & -24.31\%          & -19.81\%          & -34.69\%          \\ \hline
\textbf{UoMo w/o User}     & 0.1758            & 0.2464            & 0.1507             & 0.2363            & 0.0636            & 0.2301            \\ \hline
$\Delta$        & -88.38\%          & -67.47\%          & -80.36\%           & -63.13\%          & -89.61\%          & -76.53\%          \\ \hline
\textbf{Pre-train}    & 0.1853            & 0.2585            & 0.1635             & 0.2457            & 0.0668            & 0.2324            \\ \hline
\end{tabular}
}
}
\end{table}

\subsection{Scaling performance of UoMo}
\label{scaling_sec}

\begin{figure}[t]
    \centering
    \subfigure[Scaling performance on different tasks.]
    {\includegraphics[width=0.47\linewidth]{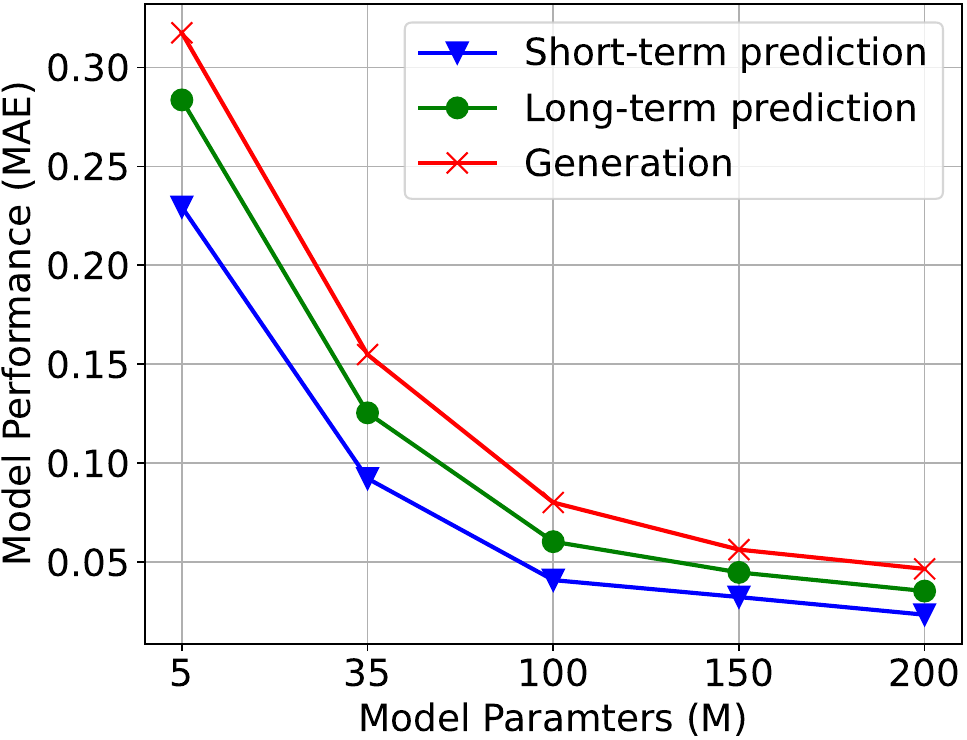}}
    \quad
    \subfigure[Scaling relationship between model parameters and data size.]{\includegraphics[width=0.47\linewidth]{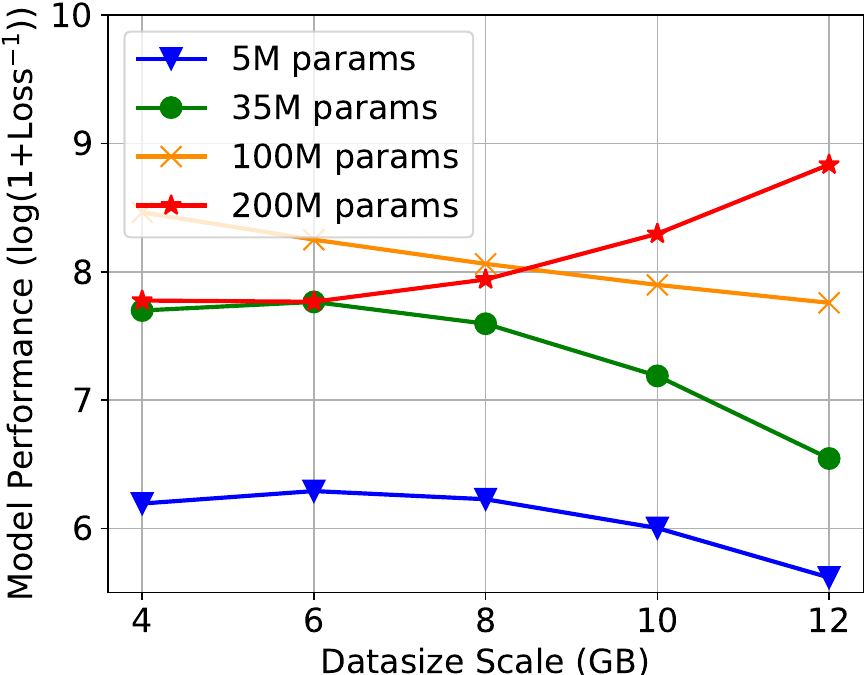}}
    \caption{Scaling performance of UoMo.}
    \label{scale2}
\end{figure}

Scaling performance reflects the relationship between model parameters, data size, and overall performance. Understanding the scaling performance of foundation models provides valuable guidance for parameter selection during model deployment, which optimizes computational and storage overheads for the entire system. 
We explore the relationship between model size and performance across tasks, as shown in Figure~\ref{scale2}(a). Smaller models improve quickly with more parameters, while larger models show diminishing returns.
We attribute this to parameter redundancy in larger models relative to fixed training data, leading to overfitting and limiting performance improvements.
To assess scaling performance, we evaluate the model with varying dataset sizes in Figure~\ref{scale2}(b). Larger models degrade with small datasets, but as data volume increases, they leverage their extensive parameters to improve performance. In contrast, smaller models struggle with diverse features, leading to performance drops.
From these observations, we identify a scaling regulation for UoMo: \emph{Simply increasing model parameters does not guarantee better performance in mobile traffic forecasting.} The optimal model size depends on the available data, prompting further investigation into the relationship between model scale and factors like urban size, population, and temporal granularity to enhance performance.

\section{Deployment Applications}

\emph{Deployment}. To validate UoMo's prediction effectiveness, we deploy the model on the Jiutian platform, an AI platform developed by China Mobile, featuring functions like scenario construction, network simulation, optimization strategy formulation, and performance evaluation.
The Jiutian platform offers full-element network simulation capabilities, enabling efficient simulation of interactions between communication systems and user behavior. It also supports operators in developing customized algorithms and applications, deploying them into production environments, and performing product validation and testing with real network data. The platform has now been fully deployed within China Mobile, supporting network development across 31 provinces in China.
We select predefined urban layout and human mobility data. UoMo is deployed in the mobile traffic module, with its predictions feeding into the optimization selection module.
We focus on 3 optimization scenarios (highlighted in yellow in the figure) and evaluate the performance via network coverage, throughput, and energy consumption. 
The training of UoMo is conducted on 4 NVIDIA A100 GPUs (80GB each) using PyTorch 2.0.1. Table~\ref{running} summarizes the model’s parameters and training/inference time per sample.

\begin{table}[h]
\caption{Model deployment efficiency on JiuTian. The training/inference time refers to the unit time per sample, which is obtained by dividing the time taken to generate a batch of data by the number of samples in the batch.}
\label{running}
{
\resizebox{0.4\textwidth}{!}{
\begin{tabular}{c|c|c|c|c|c}
\hline
\multirow{2}{*}{\textbf{Model}} & \multirow{2}{*}{\textbf{Layers}} & \textbf{Hidden}      & \textbf{Parameter}      & \textbf{Training}                  & \textbf{Inference} \\
                                &                                  & \textbf{feature}     & \textbf{scale}          & \textbf{time}                      & \textbf{time}      \\ \hline
\textbf{Open-Diff}                         & 6                               & 128                  & 10M                    & 0.09 min                           & 0.021 min           \\ \hline
\textbf{UniST}                           & 12                               & 512                  & 30M                   & 0.19 min                           & 0.009 min                  \\ \hline
\textbf{Lagllama}                        & 8                                & 144                  & 10M                    & 0.21 min                           & 0.011min            \\ \hline
\multirow{3}{*}{\textbf{UoMo}}           & 12            & 128 & 5M   & 0.24 min                          & 0.043 min                                  \\ \cline{2-6} 
                                & 16             & 256 & 35M & 0.32 min                          & 0.054 min       \\ \cline{2-6} 
                                & 20            & 768 & 200M & 0.82 min                                  & 0.162 min
                \\ \hline
\end{tabular}
}
}
\end{table}

\begin{figure}[ht]
  \centering
  \includegraphics[width=0.7\linewidth]{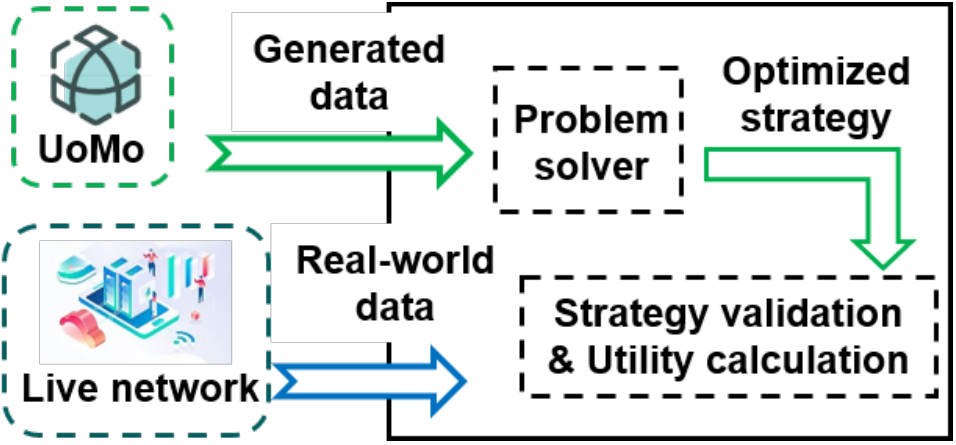}
  \caption{Optimization workflow via UoMo.}
  \label{opworkflow}
\end{figure}

\emph{Optimization}.
Our optimization method on the Jiutian platform is shown in figure~\ref{opworkflow}. First, we use UoMo or other mobile traffic forecasting algorithms to generate traffic data. The platform then uses this generated data to formulate and solve the problem of network optimization and planning. After obtaining the optimal network strategy, we input real live network traffic data into the platform to validate the optimization strategy and assess its performance.

\subsection{BS deployment}
\label{app_bs_dep}

\begin{figure}[t]
    \centering
    \subfigure[BS deployment.]
    {\includegraphics[width=0.47\linewidth]{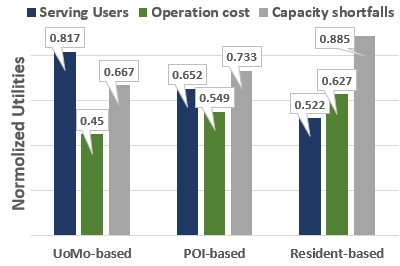}}
    \subfigure[BS sleep control optimization.]{\includegraphics[width=0.49\linewidth]{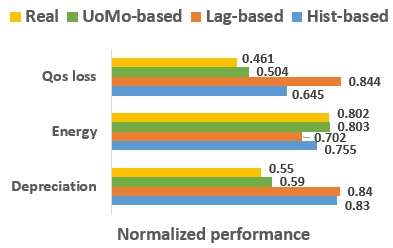}}
    \caption{Optimization results with live system data.}
    \label{app}
\end{figure}

We examine a BS deployment planning at the grid level within a discrete-time $\{1,...t\}$ framework. The target area consists of $N$ grids, and $M$ BSs need to be deployed across the grids.
Each grid experiences varying mobile users at different times $t$, and each BS has a maximum capacity of $C$ users at any given time.
The deployment strategy's objective is to optimize the serving user count in each grid while minimizing operation cost (deploying fewer BSs in grids with fewer mobile users) and reducing capacity shortfalls (preventing situations where grids lack BSs and cannot accommodate excess users).
Since there is no historical data on mobile user distribution for the target area, we rely on forecasting methods to estimate future mobile traffic demands. Three different estimation methods are utilized: the mobile traffic generation approach (UoMo-based) using the generation mask in (\ref{generation_mask}), POI distribution (POI-based), and residential population distribution (Resident-based).
Our focus is on addressing the BS deployment problem with a long-term perspective. The decision variables include $x_i$, which denotes the number of BSs deployed in grid $i$, and $u^t_i$, which represents the mobile traffic served by grid $i$ at time $t$. Then the BS deployment problem yields

\begin{equation}
\label{app1_eq}
\begin{aligned}
    & max \sum_t^T\sum_i^N \Big( y^t_i - \beta (\hat{U}^t_i-y^t_i)^+-\alpha\sum_i^M x_i \Big )\\
    & s.t. \quad \sum_i^N x_i =M, \ \ 0 \leq y^t_i \leq min\{x_iC_0, \hat{F}^t_i\},
\end{aligned}
\end{equation}
where $(y)^+ = max(0, y)$ denote choosing the maximize value between $y$ and 0, and $\hat{U}^t_i$ is the mobile network demand that is estimated by UoMo, POI, and residential distribution.
We employ the Pulp library to solve the optimization problem described above, and we can derive 3 distinct BS deployment strategies by referring to the three estimation methods. Subsequently, we test the optimization strategies using real user mobility trajectory data in Nanchang, streamed by the Jiutian platform, with the test results presented in Figure~\ref{app}(a).
The UoMo-based strategy maximizes service revenue while significantly lowering operational costs and capacity deficits, increasing the served user ratio by 25.3\% (0.652 to 0.817), and reducing costs and deficits by 18.03\% (0.549 to 0.45) and 9.00\% (0.733 to 0.667), respectively.
This success is attributed to the UoMo takes into account the time-varying nature of network usage across different regions. By accurately estimating the mobile traffic predictions with UoMo, we can effectively capture the dynamic patterns of human activities over time in a region.
In contrast, the POI-based and resident-based strategies rely solely on static attributes to guide BS deployment. These methods are less effective in capturing the dynamic activity patterns of humans over time, and fall short when compared to those based on traffic.

\subsection{BS sleep control} 
\label{app_bs_slp}

We consider the C-RAN scenario, where the BS achieves cell coverage by activating different numbers of RRUs~\cite{2023ArtificialIF}.
BS sleep strategy involves controlling the RRU operational status (activated or sleep) based on the network load. We model the problem from the perspectives of service quality, depreciation cost, and energy consumption.
We set $N$ BSs, each of which has $M$ cells to serve at time $t$. Define the traffic load that a single RRU can serve as $c_0$, then for BS $n$, its Quality of Service (QoS) equals $Q(n) =\sum_t \sum_m max(L_{m,t}-x_{m,t}c_0, 0) /L_{m,t}$, where $L_{t,m}$ is the actual cell load and $x_{m,t}$ is the activated RRUs. Moreover, frequent switching of RRUs can lead to a reduction in the lifespan of BSs, and the depreciation yields $W(n) =\sum_t \sum_m |x_{m,t}-x_{m,t-1}|$. BS energy consumption
equals $E(n) =\sum_t \sum_m \mathcal{P}[min (L_{m,t}, x_{m,t}c_0)]$ that is determined by the load at the RRU~\cite{10.1145/3139958.3140053}, where $\mathcal{P}[L] = \alpha L+ \beta (L/c_0)$ is the energy consumption function. Therefore, the optimization objective yields:
\begin{equation}
    min \sum\nolimits_n Y(x^{(n)}_{m,t} | L^{(n)}_{m, t} )= Q(n)+ W(n)+ E(n).
\end{equation}
For the BS sleeping strategy, frequently adjusting RRUs over a short period is impractical. A more reasonable approach is to assess network dynamics over a longer period and develop a long-term adjustment strategy. Therefore, we leverage UoMo's long-term prediction capability to estimate $L_{m,t}$ via the long-term prediction mask in (\ref{short_long_pre_mask}).
As shown in Figure~\ref{app}(b), the UoMo-based strategy closely aligns with the Real-based one, achieving a 21.9\% QoS improvement (0.645 to 0.504) and up to 40.7\% lower BS depreciation (0.83 to 0.59) than the others. Although energy consumption is higher than the baselines, it matches the Real-based strategy, demonstrating UoMo’s accurate traffic prediction and strong demand alignment.

\section{Related Work}
\label{sec:rela}

\emph{Mobile traffic forecasting.} It can be 
 broadly categorized into two types: prediction and generation. Mobile traffic prediction involves estimating future values using historical data, while generation learns the underlying distribution of mobile traffic relying on external contextual information and samples new data from this distribution. Early forecasting used statistical approaches or simulation techniques~\cite{8677270, 10.1145/3321349.3321351}, but these methods typically struggled to capture complex traffic patterns. 
With the rise of machine learning, many studies used AI for mobile traffic forecasting.
For mobile traffic prediction, 
LSTM models have been used to capture long-term dependencies in traffic patterns~\cite{10.1145/3291533.3291540, 10.1145/3430199.3430208, 10.1145/3433548}.
Some studies incorporated spatial attributes into traffic prediction, with Li~\emph{et al}~\cite{10.1145/3586164} combining transformer and GCN to capture spatio-temporal correlations. Wu~\emph{et al}~\cite{9662277} combined GAN with GCN to capture spatial correlations across multiple cities. The MVSTGN model~\cite{9625773} divided urban spaces into multi-attribute graphs to capture mobile traffic features in latent space.
For mobile traffic generation, early work used GANs to capture the overall distribution of mobile traffic~\cite{Ring_2019, 10.1145/3419394.3423643}. SpectraGAN~\cite{10.1145/3485983.3494844} viewed cities as images, extracting POI and land use information via CNNs and incorporating it into traffic generation. Sun~\emph{et al}~\cite{9858874} added user usage features to a GAN network, improving the accuracy of traffic generation. Hui~\emph{et al}~\cite{10.1145/3580305.3599853} built a city knowledge graph incorporating extensive semantic features into traffic generation models. Open-Diff~\cite{10.1145/3637528.3671544} used a diffusion model to generate grid-level traffic via open contextual data.

\emph{Universal and foundation models.} These models show to excel in multitasking and zero/few-shot learning, and have been applied across various specialized domains.
Yang~\emph{et al}~\cite{yang2023fingpt} and  Zhang~\emph{et al}~\cite{Zhang2024TowardsUG} proposed foundation models aim to achieve various specialized tasks like investment, quantification, and urban navigation.
Notably, many universal models for spatio-temporal forecasting have been proposed.
Leveraging existing LLM, TEMPO~\cite{cao2024tempopromtrained}  and Time-LLM~\cite{jin2024timellmtimeseriesforecasting} introduces a prompt mechanism in the pre-trained LLM for long-term forecasting by aligning features between mobile traffic and natural language tokens with a reprogramming approach.
Some methods do not rely on existing language models but reconstruct spatio-temporal foundation models using transformer architectures.
LagLLama~\cite{rasul2024lagllamafoundationmodelsprobabilistic} used lag indices to annotate multi-dimensional periodic features such as monthly, daily, and hourly periods. TimeGPT~\cite{garza2023timegpt1} replaced the Feedforward layer in the transformer with a CNN to enhance temporal correlations. UniST~\cite{10.1145/3637528.3671662} achieved spatio-temporal prediction in urban contexts by employing a memory network.
While existing models offer advantages in forecasting mobile traffic, they are often designed for specific tasks like short-term prediction or generation~\cite{ijcai2024p0921, 10.24963/ijcai.2024/895, zhang2025largelanguagemodelsmobility}. In real-world deployments, network optimization involves multiple forecasting tasks across cities, requiring frequent model switching and complex adaptations, which increases deployment costs.
The main challenge lies in mastering various forecasting tasks while integrating contextual features such as human dynamics and geographical traits. This integration is essential for building a robust model that captures the intrinsic correlations between the environment, users, and mobile traffic.

\section{Conclusion}
\label{sec:con}

In this paper, we propose UoMo, a universal model with diffusion models for mobile traffic forecasting.
To the best of our knowledge, it is the first universal model in mobile networks that simultaneously supports diverse forecasting tasks including short-term/long-term predictions and generation. 
By capturing the temporal, spatial, human dynamics, and geographical features related to mobile traffic, UoMo exhibits robust multi-task adaptability and zero/few-shot learning capability for diverse tasks across multiple cities,
which exhibits good universality.
Moreover, we identify the scaling properties of UoMo by examining the model performance with diverse parameter scales and data sizes.
We deploy UoMo on the Jiutian platform, where it is used to optimize various aspects of network coverage, throughput, and energy consumption based on its accurate traffic forecasts.
Currently, UoMo has been implemented in Nanning, Guangxi Province, China, for real-time traffic forecasting and data streaming. The model has the potential for large-scale deployment and can effectively assist operators in designing pricing strategies and network expansion, which improves both user experience and economic revenue.

\section*{Acknowledgement}
This research has been supported in part by the National Key Research and Development Program of China under Grant 2023YFB2904801, the National Natural Science Foundation of China under Grants U23B2030, and the China Postdoctoral Science Foundation under Grant 2023M742010.

\clearpage
\printbibliography

\appendix

\section{APPENDIX}

\subsection{Description of dataset}
\label{datacollect_des}

We collected mobile traffic and user data from 9 different cities, covering over 30,000 base stations, with time granularities ranging from 15 minutes to 1 hour. POI data was crawled from the map service and includes 15 categories, such as lifestyle, entertainment, work, and dining, as shown in Table~\ref{data_des}.

\begin{table}[ht]
\caption{Description of datasets}
\label{data_des}
\renewcommand{\arraystretch}{1.1}
{
\resizebox{0.5\textwidth}{!}{
\begin{tabular}{|c|ccccc|}
\hline
\multirow{2}{*}{\textbf{Dataset}} & \multicolumn{1}{c|}{\multirow{2}{*}{\textbf{Usage}}}                                           & \multicolumn{1}{c|}{\textbf{Data}}                               & \multicolumn{1}{c|}{\textbf{Mobile}}  & \multicolumn{1}{c|}{\textbf{Mobile}} & \multicolumn{1}{c|}{\textbf{Time}}        \\
                                  & \multicolumn{1}{c|}{}                                                                          & \multicolumn{1}{c|}{\textbf{description}}                        & \multicolumn{1}{c|}{\textbf{traffic}} & \multicolumn{1}{c|}{\textbf{users}}  & \multicolumn{1}{c|}{\textbf{granularity}} \\ \hline
\textbf{Beijing}                  & \multicolumn{1}{c|}{}                                                                          & \multicolumn{1}{c|}{5G data, October, 2021, 4000+ BSs}           & \multicolumn{1}{c|}{$\checkmark$}     & \multicolumn{1}{c|}{$\checkmark$}    & 1 hour                                    \\ \cline{1-1} \cline{3-6} 
\textbf{Shanghai}                 & \multicolumn{1}{c|}{}                                                                          & \multicolumn{1}{c|}{4G data, August, 2014, 5000+ BSs}            & \multicolumn{1}{c|}{$\checkmark$}     & \multicolumn{1}{c|}{$\checkmark$}    & 1 hour                                      \\ \cline{1-1} \cline{3-6} 
\textbf{Nanjing}                  & \multicolumn{1}{c|}{\multirow{3}{*}{\begin{tabular}[c]{@{}c@{}}Model\\ training\end{tabular}}} & \multicolumn{1}{c|}{5G data, February to March, 2021, 6000+ BSs} & \multicolumn{1}{c|}{$\checkmark$}     & \multicolumn{1}{c|}{$\checkmark$}    & 15 min                                    \\ \cline{1-1} \cline{3-6} 
\textbf{Nanjing-4G}               & \multicolumn{1}{c|}{}                                                                          & \multicolumn{1}{c|}{4G data, February to March, 2021, 6000+ BSs} & \multicolumn{1}{c|}{$\checkmark$}     & \multicolumn{1}{c|}{$\checkmark$}    & 15 min                                    \\ \cline{1-1} \cline{3-6} 
\textbf{Nanchang}                 & \multicolumn{1}{c|}{}                                                                          & \multicolumn{1}{c|}{5G data, July, 2023, 5000+ BSs}              & \multicolumn{1}{c|}{$\checkmark$}     & \multicolumn{1}{c|}{$\checkmark$}    & 30 min                                    \\ \cline{1-1} \cline{3-6} 
\textbf{Nanchang-4G}              & \multicolumn{1}{c|}{}                                                                          & \multicolumn{1}{c|}{4G data, July, 2023, 7000+ BSs}              & \multicolumn{1}{c|}{$\checkmark$}     & \multicolumn{1}{c|}{$\checkmark$}    & 30 min                                    \\ \cline{1-1} \cline{3-6} 
\textbf{Shandong}                 & \multicolumn{1}{c|}{}                                                                          & \multicolumn{1}{c|}{5G data, February, 2024, 1000+ BSs}          & \multicolumn{1}{c|}{$\checkmark$}     & \multicolumn{1}{c|}{$\checkmark$}    & 1 hour                                    \\ \hline
\textbf{Hangzhou}                 & \multicolumn{1}{c|}{Zero/Few}                                                                  & \multicolumn{1}{c|}{5G data, July, 2023, 1000+ BSs}              & \multicolumn{1}{c|}{$\checkmark$}     & \multicolumn{1}{c|}{$\checkmark$}    & 1 hour                                    \\ \cline{1-1} \cline{3-6} 
\textbf{Munich}                   & \multicolumn{1}{c|}{shot tests}                                                                & \multicolumn{1}{c|}{4G data, 2022, 2500+ grid-data}                      & \multicolumn{1}{c|}{$\checkmark$}     & \multicolumn{1}{c|}{---}             & 1 hour                                    \\ \hline
\multirow{3}{*}{POI}              & \multicolumn{4}{c}{Shopping, Enterprise, Restaurant, Local Living, Transportation,}                                                                                                                                                              &                                           \\
                                  & \multicolumn{4}{c}{Public Health, Automobile, Physical facilities, Accomodation, Finance,}                                                                                                                                                       &                                           \\
                                  & \multicolumn{4}{c}{Government organs, Education, Business, Public facilities, scenic spot.}                                                                                                                                                      &                                           \\ \hline
\end{tabular}
}
}
\end{table}

\subsection{Description of baselines}
\label{bsl_des}

Statistical models. Historical moving average method (\textbf{HA}) and \textbf{ARIMA} method that integrate autoregression with average moving.
Natural language-based model.
\textbf{Time-LLM} describes time series features using natural language and uses these descriptions as prompts into a natural language pre-trained model (LLAMA-7B) for forecasting.
\textbf{Tempo} designs temporal prompts with trend and seasonal features for pre-trained models (GPT-2) to predict time series.
Spatio-temporal-based models.
\textbf{TimeGPT} replaces the Feedforward layer in the transformer with a CNN network and is trained on vast spatio-temporal data. 
\textbf{Lagllama} uses a set of lag indices to capture different periodic correlations in the time series.
\textbf{CSDI} is a conditional diffusion model that uses a masking method for time-series data forecasting and imputation.
\textbf{PatchTST} decomposes time series into multiple segments and uses transformers for feature extraction.
\textbf{UniST} segments spatio-temporal data and fine-tunes the model using geographical proximity and temporal correlations.
Dedicated models for mobile traffic forecasting.
\textbf{SpectraGAN} converts mobile traffic generation into an image generation problem and utilizes a CNN-based GAN network for traffic forecasting.
\textbf{KEGAN} is a hierarchical GAN that utilizes a self-constructed Urban Knowledge Graph (UKG) to explicitly incorporate urban features during the forecasting process.
\textbf{ADAPTIVE} leverages the UKG and a BS aligning scheme to transfer mobile traffic knowledge from one city to another.
\textbf{Open-Diff} utilizes open contextual data like satellite images, residential count, and POI distribution to generate mobile traffic data.

\subsection{Proof of Lemma 1}
\label{proof_lemma}

The forward chain of DDPM gradually adds Gaussian noise $\epsilon \sim N(0, 1)$ to the original data as $q(x_k |x_{k-1}) = N(\sqrt{1-\beta_k} x_{k-1}, \beta_k \textbf{I})$, $\{\beta_k \in (0,1), k\in (1,K)\}$ is a set of scheduled noise weight, and the generated noisy data in step \emph{k} can be calculated by $x_k = \sqrt{\hat{\alpha}_k}x_0 + (1-\hat{\alpha}_k) \epsilon.$ The reversed chain utilizes a denoising network $p_\theta$ to recurrently recover $x_K$ to original data $x_0$ that yields  $p_\theta(x_{k-1}|x_k) =N(\mu_\theta(x^k, k), \sigma_\theta(x_k, k)\textbf{I})$. 
The objective of the diffusion model is essentially to maximize the log-likelihood function of the denoising network $p_\theta$ for the initial data $x_0$, 
\emph{i.e.,} 
\begin{equation}
    L(\theta) = \mathbb{E}_{x_0 \sim q(x_0)} \{-logp_\theta(x_0)\}. 
\end{equation}
Subsequently, this function is optimized using the Variational Lower Bound (VLB), which can be expressed as:
\begin{equation}
\begin{aligned}
       & -log p_\theta (x_0) \leq -log p_\theta (x_0) + D_{KL}(q_(x_{1:T}|x_0) || p_{\theta}(x_{1:T}|x_0)) \\
       & \quad \quad   = \mathbb{E}_{q_(x_{1:T}|x_0)} \{log \frac{q_(x_{1:T}|x_0)}{p_{\theta}(x_{0:T}|x_0))} \}.
\end{aligned}
\end{equation}
Taking the expectation on both sides of the above equation and applying Fubini's theorem~\cite{10.1145/3626235}, we can derive:
\begin{equation}
\begin{aligned}
    L(\theta) = &\mathbb{E}_{q(x_0)} \{-logp_\theta(x_0)\} \leq \mathbb{E}_{q(x_0)} \bigg \{ \mathbb{E}_{q_(x_{1:T}|x_0)} \{log \frac{q_(x_{1:T}|x_0)}{p_{\theta}(x_{0:T}|x_0))} \} \bigg \} \\
    & \quad \quad  = \mathbb{E}_{q_(x_{0:T})} \{log \frac{q_(x_{1:T}|x_0)}{p_{\theta}(x_{0:T}|x_0))} \}  \triangleq L_{vb}(\theta).
\end{aligned}
\label{lvb_loss}
\end{equation}
We can minimize the upper bound of $L(\theta)$ by minimizing $L_{vb}$, thereby maximizing the log-likelihood function of $p_{\theta}$.
Ho~\emph{et al.} \cite{10.5555/3495724.3496298}
proved that $L_{vb}(\theta)$ can be further parameterized by $ \mu_\theta(x_k, k)=\alpha_k^{-0.5} [x_k-\beta_k (1-\hat{\alpha}_k)^{-0.5}\epsilon_\theta(x_k,k)]$, and $\sigma_\theta$ can be parameterized as $\sigma_\theta(x_k, k) = \sqrt{(1-\hat{\alpha}_{k-1}) / (1-\hat{\alpha}_{k})  \beta_k}$. The network $p_\theta$ can then be optimized by the following objective:
\begin{equation}
    \underset{\theta}{min} L_{vb}(\theta) \approx  \underset{\theta}{min} \mathbb{E}_{x_0 \sim q(x_0), \epsilon \sim N(0,I)} [||\epsilon - \epsilon_\theta(x_k,k)||_2^2].
    \label{obj_noise}
\end{equation}
The objective in equation~(\ref{obj_noise}) is fundamentally equivalent to that of InfoNCE in contrastive learning. We use $p_{\theta}$ to represent the probability in the mutual information as
$I(e,y) = p_{\theta}(e_{0:K}|y)/ p_{\theta}(e_{0:K})$. In this way, the origin InfoNCE loss can be rewritten as:
\begin{equation}
\begin{aligned}
    &     L = \underset{e \in \mathbb{B}}{\mathbb{E}}-log\frac{p_{\theta}(e_{0:K}|y)/ p_{\theta}(e_{0:K})}{p_{\theta}(e_{0:K}|y)/ p_{\theta}(e_{0:K})+\sum_{e'} p_{\theta}(e'_{0:K}|y)/ p_{\theta}(e'_{0:K})} \\
    & \ \ \ = \underset{e \in \mathbb{B}}{\mathbb{E}}log \{ 1+\frac{p_{\theta}(e_{0:K})}{p_{\theta}(e_{0:K}|y)} \cdot N \mathbb{E}_{e'}\frac{p_{\theta}(e'_{0:K}|y)}{p_{\theta}(e'_{0:K})} \}, \\
\end{aligned} 
\end{equation}
where $e'$ denotes all negative samples.
Referencing the parameterization in \cite{NEURIPS2021_958c5305} where $p_{\theta}(x_{k-1}|x_k) = \sum_{x_0 \in q} q(x_{k-1|x_k, x_0})p_{\theta}(x_0|x_k)$, the above loss can be further formulated as:
\begin{equation}
\label{cl_loss}
\begin{aligned}
    & L \approx  \underset{\mathbb{B}}{\mathbb{E}} \bigg \{ \mathbb{E}_q \{ -log \frac{p_{\theta}(e_{0:K}|y)}{q(e_{1:K}|e_0)} \} - logN\mathbb{E}_{e'}\mathbb{E}_{q} \{ -log \frac{p_{\theta}(e'_{0:K}|y)}{q(e'_{1:K}|e'_0)} \} \bigg \}  \\
    & \ \ = L^e_{vb} - logN\sum\nolimits_{e'} L^{e'}_{vb} \\
    & \ \ \doteq \mathbb{E} \bigg \{ (\Vert \epsilon -\epsilon_{\theta}(e, k | y) \Vert^2 - \lambda \sum\nolimits_{e'} \Vert \epsilon -\epsilon_{\theta}(e',k | y) \Vert^2) \odot m  \bigg \},
\end{aligned}    
\end{equation}
where symbol $\doteq$ denotes the loss function we used during the model training process and $\lambda$ is a scaling parameter proportional to $logN$.

\end{document}